\documentclass{article}

\PassOptionsToPackage{numbers, compress}{natbib}
\usepackage[preprint]{neurips_2026}

\usepackage[utf8]{inputenc}
\usepackage[T1]{fontenc}
\usepackage{hyperref}
\usepackage{url}
\usepackage{microtype}
\usepackage{marvosym}
\usepackage{graphicx}
\usepackage{subfigure}
\usepackage{multirow}
\usepackage{makecell}
\usepackage{booktabs}
\usepackage{placeins}
\usepackage{threeparttable}
\usepackage{bm}
\usepackage{array}
\usepackage[table]{xcolor}
\usepackage{listings}
\usepackage{enumitem}
\usepackage[most]{tcolorbox}
\usepackage{amsmath}
\usepackage{amssymb}
\usepackage{amsfonts}
\usepackage{mathtools}
\usepackage{amsthm}
\usepackage{nicefrac}
\usepackage[capitalize, noabbrev]{cleveref}
\usepackage[textsize=tiny]{todonotes}

\graphicspath{{./}}

\theoremstyle{plain}

\theoremstyle{definition}

\theoremstyle{remark}

\newtcolorbox{promptbox}[1]{
  breakable,
  colback=gray!10,
  colframe=gray!50,
  coltitle=black,
  fonttitle=\bfseries\small,
  title=#1,
  boxrule=0.5pt,
  arc=2pt,
  left=6pt,
  right=6pt,
  top=4pt,
  bottom=4pt
}

\title{Joint Reward Modeling: Teaching Reward Models to Judge Like Experts}

\makeatletter
\newcommand{\blfootnotetext}[1]{%
  \begingroup
    \renewcommand{\thefootnote}{}%
    \renewcommand{\@makefnmark}{}%
    \long\def\@makefntext##1{\noindent##1}%
    \footnotetext{#1}%
  \endgroup
}
\renewcommand{\@noticestring}{}
\makeatother

\author{%
  \begin{minipage}[t]{\textwidth}\centering
    Yankai Yang\textsuperscript{*\,1,2} \quad
    Yancheng Long\textsuperscript{*\,1,2} \quad
    Hongyang Wei\textsuperscript{2} \quad
    Wei Chen\textsuperscript{2} \quad
    Kaiyu Jiang\textsuperscript{2} \\[3pt]
    Changyi Liu\textsuperscript{2} \quad
    Jiankang Chen\textsuperscript{2} \quad
    Tianke Zhang\textsuperscript{2} \quad
    Haonan Fan\textsuperscript{2} \quad
    Kaiyu Tang\textsuperscript{2} \\[3pt]
    Bin Wen\textsuperscript{\textdagger\,\Letter\,2} \quad
    Fan Yang\textsuperscript{2} \quad
    Tingting Gao\textsuperscript{2} \quad
    Han Li\textsuperscript{2} \quad
    Shuo Yang\textsuperscript{\Letter\,1}
  \end{minipage}%
}

\begin{document}

\maketitle
\blfootnotetext{%
  \textsuperscript{*}Equal contribution.\quad
  \textsuperscript{\textdagger}Project leader.\quad
  \textsuperscript{\Letter}Corresponding authors.\\[1pt]
  \textsuperscript{1}Harbin Institute of Technology, Shenzhen.\quad
  \textsuperscript{2}Kuaishou Technology.\\[1pt]
  Correspondence to: Shuo Yang $<$shuoyang@hit.edu.cn$>$, Bin Wen $<$wenbin@kuaishou.com$>$.%
}

\begin{abstract}
Reward models are critical for reinforcement learning from human feedback. Their quality directly affects the alignment and reliability of generative models. For complex tasks such as image editing, reward models must judge local visual quality, understand editing instructions, and capture cross-region semantic relations and implicit logical constraints. Existing reward modeling methods often face a trade-off between efficiency and evaluation ability. Discriminative reward models can directly align with human preferences and provide efficient, stable scores, but they often lack fine-grained semantic analysis. Generative reward models can perform deeper evaluation through language-based reasoning, but they are costly at inference time and are hard to adapt directly to preference comparisons. We propose \textbf{Joint Reward Modeling (JRM)}, which aims to teach reward models to judge like experts. During training, JRM learns fine-grained semantic analysis from language-based evaluation supervision and aligns with human choices through preference ranking. At inference time, JRM still uses efficient discriminative scoring. In this way, JRM turns the semantic analysis learned during training into efficient reward judgment, enabling fast, stable, and expert-like evaluation. Experiments show that JRM achieves state-of-the-art performance on MMRB2 and EditReward-Bench, and significantly improves the stability and effectiveness of downstream online reinforcement learning.

\end{abstract}

\begin{figure}[!t]
\centering
\includegraphics[width=0.98\textwidth]{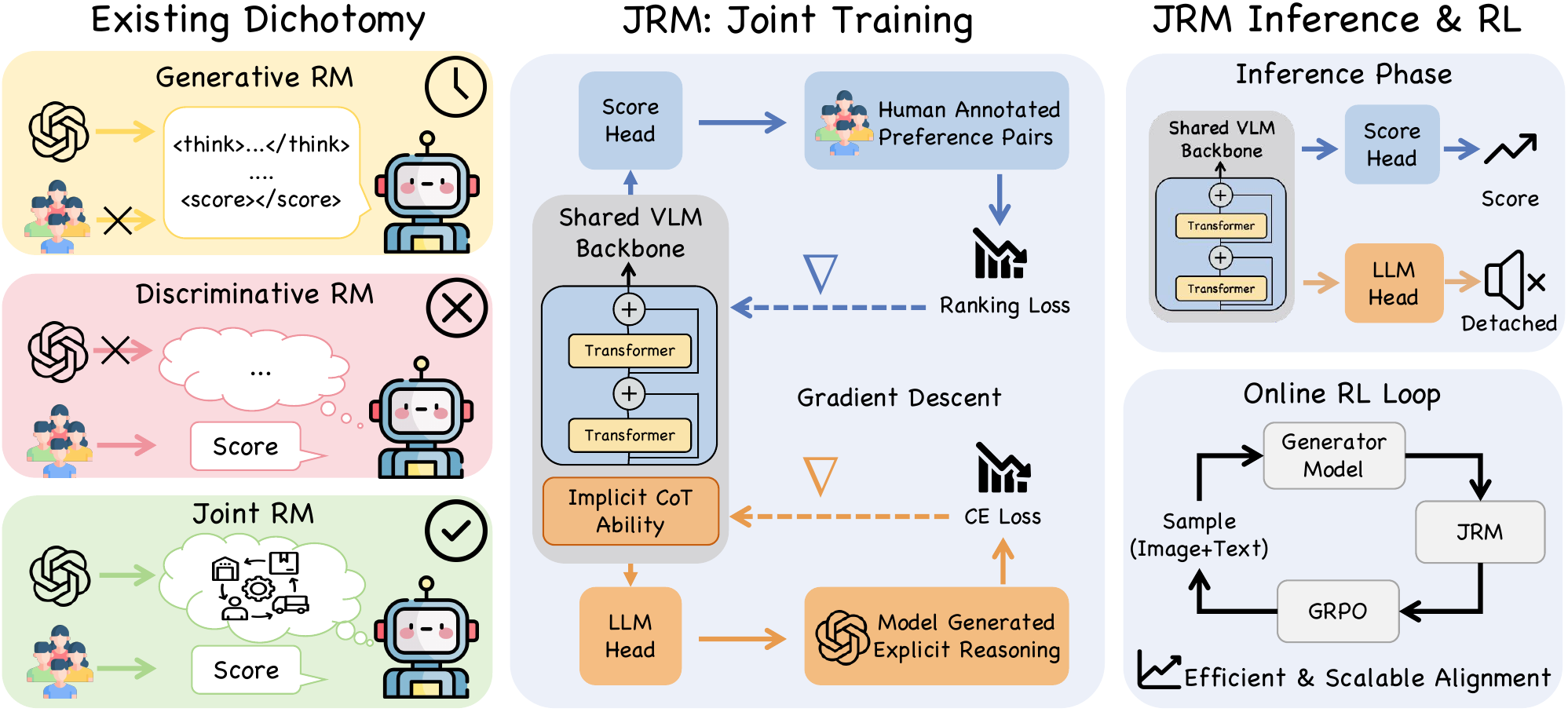}
\caption{\textbf{Overview of Joint Reward Modeling.} JRM uses language-based evaluation supervision to shape a shared reward representation. At inference time, it keeps only the efficient discriminative scoring path. This helps bridge the gap between evaluation ability and inference efficiency, and improves both reward benchmark accuracy and downstream RL alignment.}
\label{fig:overview}
\end{figure}

\section{Introduction}
\label{sec:introduction}

Reward models are a core component of reinforcement learning from human feedback (RLHF) frameworks~\cite{flowgrpo,dancegrpo}. Their primary role is to map human preferences into optimizable scalar signals, which support stable policy alignment. As multimodal foundation models evolve from perception toward reasoning and content creation, reward models are increasingly expected to handle complex generative tasks such as image editing. In these settings, reward models must go beyond surface-level similarity and capture global semantic consistency across regions and images, as well as implicit logical constraints.

In practice, existing reward modeling approaches mainly fall into two categories: discriminative and generative. Discriminative reward models~\cite{editreward,hpsv3,basereward,InternLM-XComposer} typically adopt preference learning paradigms. They directly align with human preference signals using ranking loss, which gives them stable, accurate, and efficient performance in sample ranking and quality discrimination. These models naturally leverage human preference data and provide low-latency, low-variance reward feedback in reinforcement learning loops. However, their training objective mainly constrains the final preference order. It provides limited supervision for the fine-grained semantic analysis needed in complex evaluation. As a result, discriminative reward models can still be affected by local visual patterns or shallow statistical cues when facing cross-region consistency, numerical relations, and fine-grained text-image alignment.

In contrast, generative reward models~\cite{editscore,dancereward,MM-RLHF,onereward} show stronger potential in semantic understanding and language-based evaluation. They can generate intermediate analysis text and inspect image editing results in more detail. However, their main challenges are preference modeling and inference efficiency. Generative reward models are typically trained with language modeling objectives and optimized using cross-entropy loss. Their learning signals emphasize the coherence of generated evaluation text, rather than direct and fine-grained relative preference modeling among candidate samples. In real-world settings, human feedback is often provided as preference comparisons or coarse selections. Such feedback rarely includes the detailed scores or explanations required by generative models. This mismatch makes it hard for generative reward models to align reliably and directly with human preferences. Moreover, during reinforcement learning, generative reward models must repeatedly generate intermediate evaluation text. This adds high computation cost and inference latency. Their reward signals are also expensive and often less stable than those produced by ranking-based discriminative models, especially in sample comparison tasks. These factors limit their use in large-scale online RL.

This analysis reveals a key conflict in reward modeling. Generative reward models can provide more detailed semantic analysis, but they are hard to use for efficient and stable preference modeling. Discriminative reward models can align with human preferences accurately and efficiently, but their training objective gives limited constraints on the semantic structure needed for complex evaluation. An ideal reward model should combine both advantages. It should receive fine-grained semantic analysis supervision during training, while keeping efficient and stable discriminative scoring during inference. This leads to the central question of this work:
\textbf{Can a model gain stronger semantic analysis ability without sacrificing preference modeling accuracy or inference efficiency?}

Human experts are usually not born with fast and stable judgment. They first go through extensive explicit analysis, reasoning, and feedback. Over time, they turn complex semantic relations and logical rules into efficient intuitive judgment. Inspired by this process, we argue that a reward model does not have to generate a full Chain-of-Thought at inference time. What matters more is whether training helps the model organize evaluation-related semantic structures, logical constraints, and cross-region consistency in its internal representations. In other words, explicit language-based analysis can serve as supervision during training. At inference time, the model can use a compact discriminative path to make stable judgments.

Based on this insight, we propose \textbf{Joint Reward Modeling (JRM)}. JRM is not a simple compromise between discriminative and generative approaches. It combines their strengths in one training framework (see Figure~\ref{fig:overview}). On a shared vision-language backbone, JRM jointly optimizes a ranking loss for preference learning and a cross-entropy loss for language modeling. During training, language-based evaluation supervision provides fine-grained semantic analysis signals. It helps the shared representation capture semantic relations, logical constraints, and cross-region consistency needed for high-quality evaluation. During inference, the language generation path is removed, and only the efficient discriminative scoring path is kept. This training-inference decoupling enables fast, stable, and expert-like reward evaluation without test-time text generation.

Experiments show that JRM achieves strong performance without test-time text generation: 85.1\% on EditReward-Bench~\cite{editreward} and 69.3\% on MMRB2~\cite{mmrb2}, improving over GPT-5 by 9.6 and 7.4 points, respectively. Representation analysis further shows an effective rank of 91.77, substantially higher than the baseline's 46.86. When used as the reward model for downstream Flow-GRPO optimization, JRM yields gains of \textbf{1.00} and \textbf{0.50} on GEdit-Bench and ImageEdit-Bench~\cite{geditbench,imageeditbench}, exceeding the gains from GPT-4.1 (+0.45 and +0.26).

Our main contributions are summarized as follows:
\begin{itemize}[leftmargin=1.2em]
\item \textbf{We propose Joint Reward Modeling (JRM), an efficient visual reward modeling framework for expert-like judgment.} JRM jointly optimizes preference learning and language modeling objectives over shared representations, preserves the strengths of discriminative models in preference modeling, accuracy, and inference efficiency, and uses language-based evaluation supervision to learn more semantically discriminative reward representations for complex semantic relations, global consistency, cross-region relations, and implicit logical constraints, without test-time generation overhead.
\item \textbf{We construct and will release a structured evaluation-language supervision dataset based on open-source preference data.} The dataset contains 200K explanatory supervision targets aligned one-to-one with the original preference samples, covering region grounding, instruction following, visual quality, and global consistency, and provides richer supervision for training expert-like reward models.
\item \textbf{We achieve new state-of-the-art results on multiple reward modeling benchmarks and downstream online reinforcement learning tasks.} Further ablations, control experiments, matched-backbone comparisons, and representation analyses validate JRM's effectiveness and show that its gains mainly come from structured evaluation-language supervision aligned with reward targets.
\end{itemize}

\section{Related Work}
\label{sec:related_work}

\paragraph{Reward Modeling for Generative Alignment.}
Reward models (RMs) map human or AI feedback to scalar rewards for policy optimization~\cite{christiano2017deep,ouyang2022training}. Early methods mainly use discriminative preference learning, such as pairwise or listwise ranking from comparisons~\cite{christiano2017deep,ziegler2019fine}. They are efficient and work well with PPO, DPO, and GRPO~\cite{PPO,DPO,GRPO}, but they can rely on surface patterns and miss global semantic relations.

Reward modeling has also been widely used in visual generation. CLIP-based models such as PickScore and HPSv2~\cite{PickScore,HPSv2} evaluate image-text consistency and aesthetics, while VIEScore~\cite{ku2024viescore} uses multimodal large language models for interpretable assessment. Recent unified multimodal models also support image generation and editing in one framework~\cite{xiao2025omnigen,bagel,seedream2025seedream,wu2025qwen,labs2025flux}. Image editing is harder because evaluation must compare the source and edited images. EditReward~\cite{editreward} trains a discriminative ranking model with human preferences. UniPic 2.0~\cite{unipic2} uses closed-source evaluators for online RL~\cite{flowgrpo,he2025tempflow}. EditScore~\cite{editscore} distills reasoning behaviors to reduce inference cost. SpatialReward~\cite{ConcurrentWork1} introduces explicit spatial reasoning. Still, many methods rely on holistic scoring and weakly model fine-grained semantic relations.

\paragraph{Chain-of-Thought for Reasoning and Evaluation.}
Chain-of-Thought (CoT)~\cite{CoT,ToT} improves reasoning and interpretability by generating intermediate steps. In alignment, it is used in LLM-as-a-Judge and VLM-as-a-Judge frameworks~\cite{r1reward,unified,unifiedcot,paco,criticoutloud,comanici2025gemini} for step-by-step analysis of semantic consistency and logical correctness.

However, explicit CoT generation is costly in large-scale online RL. This motivates methods that use language-based analysis during training but avoid test-time generation.

\paragraph{Training-Time Language Supervision for Efficient Evaluation.}
Recent studies show that reasoning or explanation supervision during training can improve internal representations, even when CoT text is disabled at inference time~\cite{ImplicitCoT,AdaCoT,keye1.5b}. Thus, language-based analysis can also serve as a training signal for shaping representations.

Reward modeling has not fully explored how to combine this supervision with preference ranking. JRM addresses this gap by jointly optimizing language and reward objectives. It uses language-based evaluation supervision during training while preserving efficient discriminative scoring at inference time.

\section{Methodology}
\label{sec:method}

\subsection{Training Data and Supervision Signals}

JRM is trained on data constructed from EditReward-Data~\cite{editreward}.
To provide language-based evaluation supervision, we use GPT-5 to generate score-aligned explanatory critiques for 200K samples.
The full prompts and control variants are described in Appendix~\ref{appendix:data}.
We will open-source this dataset to facilitate reproduction.

\subsection{Model Architecture and Joint Objective}

JRM adopts a shared vision-language backbone with task-specific output heads.
Given an image $x$ and instruction $c$, the backbone encoder produces a shared representation:
\begin{equation}
\mathbf{h} = E(x, c).
\end{equation}

A lightweight discriminative head maps $\mathbf{h}$ to a scalar reward score:
\begin{equation}
r = f_\theta(\mathbf{h}),
\end{equation}
which is used for preference ranking.
A conditional language head models the generation of semantic evaluations:
\begin{equation}
p(y \mid x, c) = g_\phi(\mathbf{h}).
\end{equation}

Reward learning is formulated as a preference ranking problem.
To capture potential inconsistencies in annotation, we adopt the uncertainty-aware ranking approach from HPSv3~\cite{hpsv3}.
By modeling the reward score as a Gaussian distribution $r \sim \mathcal{N}(\mu, \sigma)$, the preference probability becomes:
\begin{equation}
\begin{split}
P(x_i \succ x_j \mid c) &= \iint \text{sigmoid}(r_i - r_j) \mathcal{N}(r_i | \mu_i, \sigma_i) \\
&\quad \times \mathcal{N}(r_j | \mu_j, \sigma_j) \, dr_i \, dr_j,
\end{split}
\end{equation}
which leads to the ranking loss:
\begin{equation}
\mathcal{L}_{\text{rank}} = - \mathbb{E}_{(i,j)} \big[\log P(x_i \succ x_j \mid c)\big].
\end{equation}

The language supervision head minimizes the standard cross-entropy loss over the target explanation $y$:
\begin{equation}
\mathcal{L}_{\text{LM}} = - \sum_{t=1}^{T} \log p(y_t \mid y_{<t}, x, c).
\end{equation}
The joint training objective is:
\begin{equation}
\mathcal{L}_{\text{total}} = (1-\alpha)\,\mathcal{L}_{\text{rank}} + \alpha\,\mathcal{L}_{\text{LM}},
\end{equation}
where $\alpha$ controls the strength of language supervision ($\alpha = 0.7$ for our JRM).

\subsection{Evaluation-Related Internal Representations and Efficient Inference}

Joint training encourages the shared representation $\mathbf{h}$ to encode semantic factors that determine editing quality, such as instruction satisfaction, cross-region consistency, and implicit logical constraints.

The ranking loss requires these factors to support stable preference ordering. The language modeling loss provides structured evaluation supervision and encourages the model to organize evaluation-related semantic structures in its internal representations. As a result, the semantic analysis learned during training can be turned into reward judgment through a lightweight discriminative head:
\begin{equation}
r = f_\theta(\mathbf{h}_{\text{eval}}).
\end{equation}
Here, $\mathbf{h}_{\text{eval}}$ denotes the evaluation-related representation used for reward judgment.

Unlike explicit Chain-of-Thought methods, JRM does not generate intermediate evaluation text at inference time. Language supervision acts as a training-time constraint and helps the model avoid collapse to shallow features:
\begin{equation}
\operatorname{rank}\big(\mathrm{Cov}(\mathbf{h})\big) \uparrow \quad \text{under joint training}.
\end{equation}
We empirically validate this hypothesis in Section~\ref{sec:svd_analysis}, where joint training nearly doubles the effective rank compared to the baseline.

During inference, JRM uses only the backbone and reward head:
\begin{equation}
r = f_\theta(E(x, c)),
\end{equation}
which ensures low-latency evaluation.
The language head can be enabled for diagnostic purposes but does not affect reward computation. For training details, see Figure~\ref{fig:ablation_training} and Appendix~\ref{appendix:training}.

\section{Experiments}
\label{sec:experiments}

This section evaluates Joint Reward Modeling (JRM) systematically. We first test benchmark performance and use both language-weight ablations and language-form controls to study language supervision. We then analyze representation structure, self-diagnosis and self-correction behavior, and downstream online RL efficiency.

\subsection{Performance on Reward Modeling Benchmarks}

We evaluate JRM's performance on multiple public reward model benchmarks~\cite{editscore,mmrb2}, covering different types of image editing instructions, semantic complexity, and preference annotation formats. Comparison methods include pure discriminative reward models and generative reward models based on explicit reasoning generation.

\begin{table}[!t]
    \centering
    \scriptsize
    \setlength{\tabcolsep}{3pt}
    \caption{\textbf{Performance on reward modeling benchmarks.} Results of representative models on EditReward-Bench and MMRB2; shaded columns denote the primary aggregated metric for each benchmark.}
    \label{tab:reward_benchmarks}
    \resizebox{\textwidth}{!}{%
    \begin{tabular}{l l c c >{\columncolor{gray!18}}c c c >{\columncolor{gray!18}}c}
        \toprule
        \multirow{2}{*}{\textbf{Model}} & \multirow{2}{*}{\textbf{Type}} & \multicolumn{3}{c}{\textbf{EditReward-Bench}} & \multicolumn{3}{c}{\textbf{MMRB2 (Pointwise)}} \\
        \cmidrule(lr){3-5} \cmidrule(lr){6-8}
        & & \textbf{Prompt Following} & \textbf{Consistency} & \textbf{Overall} & \textbf{Single-Image} & \textbf{Multi-Image} & \textbf{Overall} \\
        \midrule
        \multicolumn{8}{c}{\textbf{\textit{Proprietary Models}}} \\
        \midrule
        GPT-4.1 & Closed & 0.673 & 0.602 & 0.705 & 0.547 & 0.478 & 0.535 \\
        GPT-5 & Closed & 0.777 & 0.669 & 0.755 & 0.627 & 0.584 & 0.619 \\
        Gemini-2.5-Pro & Closed & 0.703 & 0.560 & 0.722 & 0.545 & 0.483 & 0.534 \\
        Gemini-3.0-Flash & Closed & 0.717 & 0.662 & 0.769 & 0.627 & 0.596 & 0.621 \\
        \midrule
        \multicolumn{8}{c}{\textbf{\textit{Open-Source Models}}} \\
        \midrule
        Qwen3-VL-8B & Gen. & 0.419 & 0.243 & 0.562 & 0.425 & 0.393 & 0.419 \\
        EditScore-8B~\cite{editscore} & Gen. & 0.608 & 0.594 & 0.690 & 0.579 & 0.528 & 0.570 \\
        EditReward~\cite{editreward} & Disc. & 0.832 & - & 0.792 & 0.672 & 0.590 & 0.657 \\
        \midrule
        \rowcolor{blue!6}
        \textbf{JRM (Ours)} & \textbf{Joint} & \textbf{0.854} & - & \textbf{0.851} & \textbf{0.703} & \textbf{0.646} & \textbf{0.693} \\
        \bottomrule
    \end{tabular}}
\end{table}

EditScore-8B uses the Qwen3-VL-8B model reported in EditScore~\cite{editscore}, while EditReward uses the Qwen2.5-VL-7B model reported in EditReward~\cite{editreward}; Appendix~\ref{appendix:matched_backbone} reports a Qwen2.5-VL-7B JRM for a matched-backbone check.

As shown in Table~\ref{tab:reward_benchmarks}, JRM achieves significantly better performance than pure discriminative reward models on EditReward-Bench. In the Prompt Following dimension, JRM achieves 85.4\% accuracy; in the Overall metric, it reaches 85.1\%, a 5.9\% improvement over the second-best method, demonstrating strong comprehensive evaluation capability. Notably, JRM does not introduce any additional language generation process during inference, maintaining the same inference path and computational overhead as baseline discriminative models.

On the MMRB2 benchmark, JRM also demonstrates excellent performance. In Pointwise evaluation, JRM achieves the best results across all three dimensions---Single Image, Multi Image, and Overall---with improvements of 3.1\%, 5.6\%, and 3.6\% respectively compared to previous best method. These results indicate that under the same inference efficiency conditions, introducing joint training significantly improves ranking and scoring performance for reward models.

\subsection[Impact of Language Supervision Weight alpha]{Impact of Language Supervision Weight ($\alpha$)}

\begin{figure}[!t]
\centering
\includegraphics[width=\textwidth]{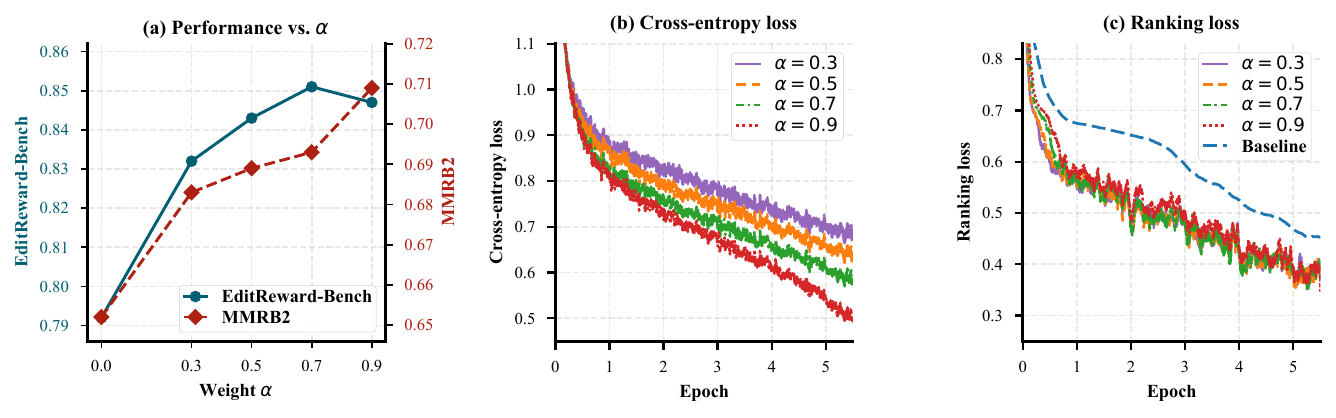}
\caption{\textbf{Ablation and training dynamics.} \textbf{(a)} Reward-model benchmark results under different language-supervision weights $\alpha$. \textbf{(b)} Cross-entropy loss for language supervision after the initial transient. \textbf{(c)} Ranking loss for reward scoring under different $\alpha$ values after the initial transient.}
\label{fig:ablation_training}
\end{figure}

To analyze the impact of language supervision, we vary the language loss weight $\alpha$ and keep all other training configurations fixed.

Figure~\ref{fig:ablation_training}(a) shows that larger $\alpha$ values bring stable improvements in accuracy and consistency across benchmarks.
The exact values corresponding to Figure~\ref{fig:ablation_training}(a) are reported in Appendix Table~\ref{tab:alpha_ablation_values}.

This indicates that language supervision is not only a regularizer, but is also related to the quality of learned representations.

Figures~\ref{fig:ablation_training}(b) and~\ref{fig:ablation_training}(c) show the training dynamics. When $\alpha > 0$, the ranking loss converges faster and more stably than the baseline ($\alpha = 0$). The cross-entropy loss also converges smoothly, suggesting that the two objectives are complementary.

\subsection{Control Experiments on Language Supervision Forms}
\label{sec:language_controls}

To further compare how different forms of language supervision affect reward learning, we construct the control experiments in Table~\ref{tab:language_controls}. All experiments use the same 200K EditReward training samples and keep the image inputs, preference labels, and reward-model architecture unchanged. Caption-only supervision uses GPT-5 to rewrite each sample into a short image-pair description. Mismatched explanations use the same pool of explanation texts but randomly shuffle their correspondence across training samples. Score-unconditioned explanations are generated without showing the original assigned score to the generation model; only the images and editing instruction are provided when generating the critique. More construction details are provided in Appendix~\ref{appendix:data}. The GEdit $\Delta$ and ImageEdit $\Delta$ columns report downstream online RL gains on the two editing benchmarks; the corresponding setup is introduced in Section~\ref{sec:online_rl}.

As shown in Table~\ref{tab:language_controls}, caption-only supervision improves over ranking-only training but remains below JRM. This suggests that descriptive language supervision brings some gains, but evaluation-style supervision is more effective for helping the shared backbone learn evaluation-related representations for reward judgment. Mismatched explanations substantially hurt performance, further showing that the gains do not come from arbitrary text supervision or random language regularization, but depend on the correspondence between explanation texts and specific samples. Score-unconditioned explanations also clearly improve over the ranking-only baseline and remain close to score-aligned JRM, suggesting that providing the original score is not a decisive factor; as long as the explanations are grounded in the images and editing instruction, they can still provide effective evaluation supervision. Overall, JRM's main benefit comes from evaluation-style language supervision aligned with specific samples and reward targets, rather than from simply adding language supervision itself.

\begin{table}[!t]
    \centering
    \footnotesize
    \setlength{\tabcolsep}{4pt}
    \caption{\textbf{Language supervision control experiments.} Comparison of ranking-only, caption-only, mismatched-explanation, score-unconditioned, and JRM training signals on reward-model benchmarks and downstream RL gains.}
    \label{tab:language_controls}
    \resizebox{\textwidth}{!}{%
    \begin{tabular}{l c c c c}
        \toprule
        \textbf{Training Signal} & \textbf{EditReward-Bench} & \textbf{MMRB2} & \textbf{GEdit $\Delta$} & \textbf{ImageEdit $\Delta$} \\
        \midrule
        Ranking-only ($\alpha=0$) & 0.792 & 0.652 & +0.82 & +0.23 \\
        Caption-only supervision & 0.823 & 0.671 & +0.89 & +0.32 \\
        Mismatched explanations & 0.683 & 0.592 & +0.49 & +0.15 \\
        Score-unconditioned explanations & 0.840 & 0.682 & -- & -- \\
        \rowcolor{blue!6}
        \textbf{JRM score-aligned explanations} & \textbf{0.851} & \textbf{0.693} & \textbf{+1.00} & \textbf{+0.50} \\
        \bottomrule
    \end{tabular}}
\end{table}

\subsection{Representation Analysis via Singular Value Decomposition}
\label{sec:svd_analysis}
\begin{figure}[!t]
\centering
\includegraphics[width=\textwidth]{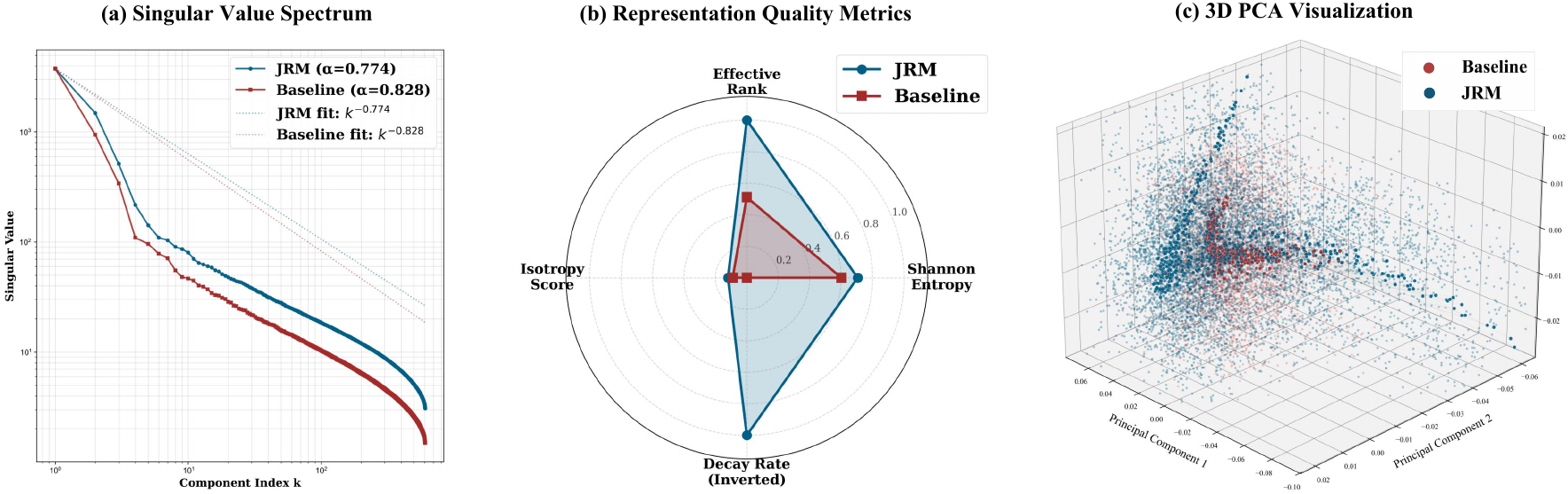}
\caption{\textbf{Representation space analysis.} \textbf{(a)} Singular value spectra of the baseline and JRM. \textbf{(b)} Representation quality metrics, including effective rank, spectral entropy, and isotropy score. \textbf{(c)} 3D PCA visualization of hidden states after Procrustes alignment. JRM shows a higher-dimensional and more dispersed representation structure across both metrics and visualization.}
\label{fig:svd_analysis}
\end{figure}

To examine whether JRM changes the internal representation structure of the reward model, we compare the ranking-only baseline ($\alpha=0$) and JRM ($\alpha=0.7$) under the same architecture and inference flow. Both models evaluate the same OmniGen2 outputs on GEdit-Bench, and we extract hidden states from corresponding shared-backbone layers. We then apply singular value decomposition (SVD) to characterize how representation energy is distributed across principal directions under different training objectives.

Results are shown in Figure~\ref{fig:svd_analysis}. JRM's singular value spectrum decays more slowly than the baseline, indicating that representation energy is less concentrated in a few dominant directions and instead spreads across more effective dimensions. Its effective rank reaches 91.77, compared with 46.86 for the baseline; spectral entropy and isotropy also improve, suggesting fuller dimensional use and a more balanced representation geometry. The PCA visualization shows the same trend: JRM representations are more dispersed, whereas baseline representations are more concentrated.

These results suggest that evaluative language supervision not only improves final ranking performance but also changes evaluation-relevant representations in the shared backbone. Table~\ref{tab:representation_control} provides a further control. Caption-only supervision improves representation diversity over the ranking-only baseline, but its effective rank, spectral entropy, and benchmark performance remain below JRM, indicating that simple descriptive language supervision cannot replace evaluative explanations aligned with the samples and reward objective.

\begin{table}[!t]
    \centering
    \footnotesize
    \setlength{\tabcolsep}{4pt}
    \caption{\textbf{Representation-space control.} Representation quality metrics and reward-model benchmark results under different training signals.}
    \label{tab:representation_control}
    \resizebox{\textwidth}{!}{%
    \begin{tabular}{l c c c c c}
        \toprule
        \textbf{Model} & \textbf{Effective Rank} & \textbf{Spectral Entropy} & \textbf{Isotropy} & \textbf{EditReward-Bench} & \textbf{MMRB2} \\
        \midrule
        Ranking-only ($\alpha=0$) & 46.86 & 3.8472 & 0.0897 & 0.792 & 0.652 \\
        Caption-only supervision & 78.67 & 4.1023 & 0.1223 & 0.823 & 0.671 \\
        \rowcolor{blue!6}
        \textbf{JRM} & \textbf{91.77} & \textbf{4.5193} & \textbf{0.1199} & \textbf{0.851} & \textbf{0.693} \\
        \bottomrule
    \end{tabular}}
\end{table}

\subsection{JRM-Guided Self-Correction Analysis}

This experiment analyzes JRM on editing results with semantic defects. We select low-VIEScore samples generated by OmniGen2~\cite{Omnigen2} on GEdit-Bench~\cite{geditbench}, use the language head to generate feedback, and use this feedback to guide correction. We focus on two complementary questions: whether JRM-guided correction improves both external VIEScore and JRM scores in a consistent direction, and whether this feedback provides actionable semantic signals across different base editors.

\begin{table}[!t]
    \centering
    \scriptsize
    \setlength{\tabcolsep}{2pt}
    \caption{\textbf{JRM-guided self-correction diagnostics.} \textbf{(a)} Correction results under different initial VIEScore thresholds. \textbf{(b)} Before/after correction results across different base editors. B. and A. denote before and after correction, respectively; shaded columns mark the primary comparison metrics.}
    \label{tab:self_correction}
    \begin{minipage}[t]{0.49\textwidth}
    \centering
    \textbf{(a) Threshold-based correction}\vspace{0.25em}

    \begin{tabular*}{\textwidth}{@{\extracolsep{\fill}}l c >{\columncolor{gray!18}}c >{\columncolor{gray!18}}c}
        \toprule
        \textbf{VIE Thr.} & \textbf{N} & \textbf{VIE $\Delta$} & \textbf{JRM $\Delta$} \\
        \midrule
        $< 7.0$ & 254 & +0.44 & +0.28 \\
        $< 5.0$ & 169 & +1.23 & +0.28 \\
        $< 3.0$ & 91 & +2.39 & +0.43 \\
        \bottomrule
    \end{tabular*}
    \end{minipage}
    \hfill
    \begin{minipage}[t]{0.49\textwidth}
    \centering
    \textbf{(b) Cross-editor diagnostic}\vspace{0.25em}

    \begin{tabular*}{\textwidth}{@{\extracolsep{\fill}}l c >{\columncolor{gray!18}}c c >{\columncolor{gray!18}}c}
        \toprule
        \textbf{Editor} & \textbf{GEdit B.} & \textbf{GEdit A.} & \textbf{ImgEdit B.} & \textbf{ImgEdit A.} \\
        \midrule
        OmniGen2 & 6.42 & 6.78 & 3.44 & 3.62 \\
        Flux-Kontext-dev & 6.31 & 6.69 & 3.60 & 3.67 \\
        Qwen-Image-Edit & 7.56 & 7.60 & 4.27 & 4.32 \\
        \bottomrule
    \end{tabular*}
    \end{minipage}
\end{table}

\begin{figure}[!t]
\centering
\subfigure[JRM-guided self-correction.\label{fig:self_correction_case}]{
    \includegraphics[height=0.34\textheight]{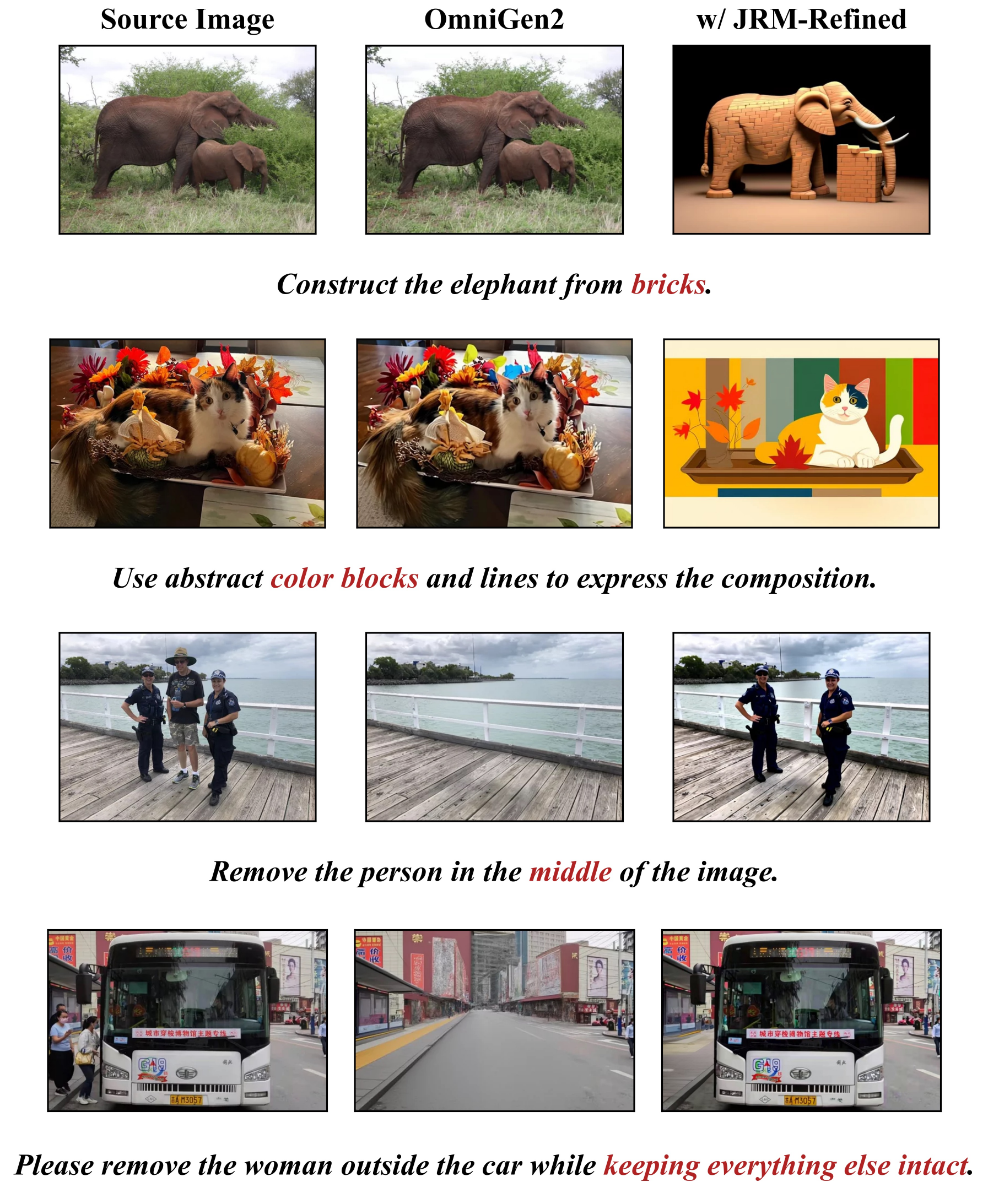}}
\hspace{0.01\textwidth}
\subfigure[RL-finetuned model comparison.\label{fig:rl_qualitative}]{
    \includegraphics[height=0.34\textheight]{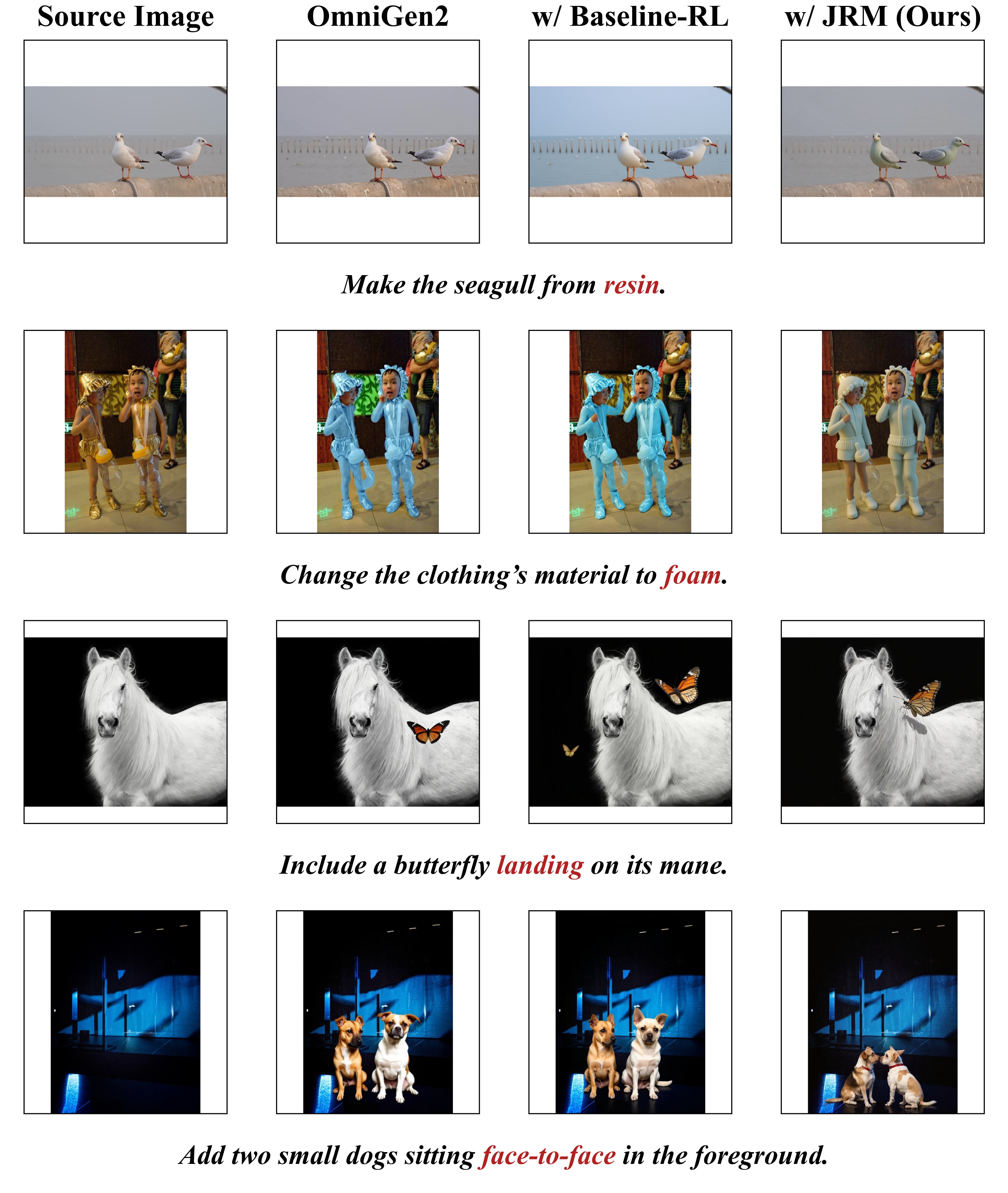}}
\caption{\textbf{Qualitative analyses.} \textbf{(a)} Example of JRM-guided self-correction. \textbf{(b)} Qualitative comparison of RL-finetuned models.}
\label{fig:qualitative_combined}
\end{figure}

As shown in Table~\ref{tab:self_correction}(a), samples with lower initial VIEScore obtain larger VIEScore gains after correction, while JRM scores also improve consistently. This indicates that the problems identified by the language head are aligned with visual-quality improvement and with the reward preference of the score head. Figure~\ref{fig:self_correction_case} shows a concrete example: the language head identifies semantic defects in the edited image and provides feedback grounded in visual evidence; guided by this feedback, the editing model generates a more semantically consistent result.

Table~\ref{tab:self_correction}(b) further shows that JRM feedback improves outputs from OmniGen2, Flux-Kontext-dev, and Qwen-Image-Edit, suggesting that this behavior is not limited to a single base editor. Overall, these results show that the language head can read out actionable semantic defects from the shared representation, and that this semantic information is consistent with final reward prediction.

\subsection{Downstream Online Reinforcement Learning}
\label{sec:online_rl}

To validate JRM's effectiveness as a reward function in practical optimization loops, we fine-tune image editing generation models under online reinforcement learning settings. Specifically, we adopt the Flow-GRPO~\cite{flowgrpo} algorithm and train the same generation model using various open-source and closed-source reward models as well as JRM as reward signals. Experiments are evaluated on two widely-used image editing benchmarks covering multiple practical editing tasks.

\begin{table}[!t]
    \centering
    \footnotesize
    \setlength{\tabcolsep}{3pt}
    \caption{\textbf{Online RL performance on GEdit-Bench and ImageEdit-Bench.} Comparison of OmniGen2 alignment with different reward signals, with additional JRM results on the UniRef-Image-Edit~\cite{wei2026uniref} backbone.}
    \label{tab:rl_results}
    \resizebox{\textwidth}{!}{%
    \begin{tabular}{l c c c >{\columncolor{gray!18}}c c >{\columncolor{gray!18}}c }
        \toprule
        \multirow{2}{*}{\textbf{Configuration}} & \multicolumn{4}{c}{\textbf{GEdit-Bench}} & \multicolumn{2}{c}{\textbf{ImageEdit-Bench}} \\
        \cmidrule(lr){2-5} \cmidrule(lr){6-7}
        & \textbf{Sem. Consistency} & \textbf{Perc. Quality} & \textbf{Overall} & \textbf{Gain $\Delta$} & \textbf{Overall} & \textbf{Gain $\Delta$} \\
        \midrule
        \multicolumn{7}{c}{\textbf{\textit{Baselines (Reported in EditScore)}}} \\
        \midrule
        OmniGen2 & 6.72 & 7.20 & 6.28 & - & 3.40 & - \\
        OmniGen2 w/ GPT-4.1 & 7.24 & 7.40 & 6.73 & +0.45 & 3.66 & +0.26 \\
        OmniGen2 w/ EditScore-8B & 7.28 & 6.89 & 6.89 & +0.61 & 3.62 & +0.22 \\
        \midrule
        \multicolumn{7}{c}{\textbf{\textit{Results in Our Environment}}} \\
        \midrule
        OmniGen2 (Reprod.) & 6.88 & 7.38 & 6.42 & - & 3.44 & - \\
        OmniGen2 w/ EditReward & 7.43 & 7.89 & 7.19 & +0.77 & 3.63 & +0.19 \\
        OmniGen2 w/ Baseline($\alpha=0$) & 7.50 & 7.91 & 7.24 & +0.82 & 3.67 & +0.23 \\
        \rowcolor{blue!6}
        \textbf{OmniGen2 w/ JRM} & \textbf{7.75} & \textbf{8.14} & \textbf{7.42} & \textbf{+1.00} & \textbf{3.94} & \textbf{+0.50} \\
        \midrule
        \multicolumn{7}{c}{\textbf{\textit{Generality on a Stronger Generator Backbone}}} \\
        \midrule
        UniRef-Image-Edit & 7.81 & 7.84 & 7.46 & - & 4.16 & - \\
        \rowcolor{blue!6}
        UniRef-Image-Edit w/ JRM & 8.08 & 7.98 & 7.58 & +0.12 & 4.24 & +0.08 \\
        \bottomrule
    \end{tabular}}
\end{table}

As shown in Table~\ref{tab:rl_results}, JRM-based RL improves both GEdit-Bench and ImageEdit-Bench~\cite{geditbench,imageeditbench}. JRM also improves UniRef-Image-Edit~\cite{wei2026uniref}, suggesting that the reward remains useful for a stronger generator.

Training curves are shown in Appendix Figure~\ref{fig:rl_training}. The JRM reward rises steadily and tracks GEdit-Bench scores well, supporting its reliability. Figure~\ref{fig:rl_qualitative} further shows that OmniGen2 aligned with JRM better handles complex instructions involving materials, textures, and spatial relations. More details and cases are provided in Appendices~\ref{appendix:training} and~\ref{appendix:rl_qualitative}.

\section{Conclusion}
\label{sec:conclusion}

This paper proposes \textbf{Joint Reward Modeling (JRM)}, an efficient visual reward modeling framework for expert-like judgment. Through joint training of language modeling and reward prediction, JRM learns fine-grained semantic analysis from language-based evaluation supervision during training. At inference time, it keeps only the discriminative scoring path. Experiments show that JRM outperforms traditional discriminative models on multiple reward model benchmarks and brings stable gains in downstream online reinforcement learning. Overall, JRM shows that semantic analysis learned during training can be turned into efficient reward judgment, providing a practical solution for scalable reward modeling and alignment.

\clearpage
\nocite{*}
\bibliographystyle{plainnat}
\bibliography{jrm_paper}

\clearpage
\appendix
\section{Language Supervision Data Construction}
\label{appendix:data}

This section provides the prompt templates used to generate structured language supervision signals for joint training and the corresponding control variants.

\textbf{Data Source and Alignment.} The language supervision data are constructed from the EditReward training split, using 200K pairwise preference samples. For each preference example, we keep the original image, edited image, instruction, and assigned score aligned one-to-one with the generated explanation target, without introducing additional images. The main explanation corpus is generated with GPT-5 using the prompts below; the auxiliary control variants are described in Appendix~\ref{appendix:control_variants}.

\subsection{Instruction Following Evaluation Prompt}

\begin{promptbox}{INSTRUCTION FOLLOWING DATA CONSTRUCTION TEMPLATE}
\small
You are an expert AI training data annotator and visual linguistics specialist.
Your task is to generate a structured evaluation entry for an image editing pair based on a pre-assigned score.

\textbf{**Inputs:**}\\
- \textbf{Instruction:} ``\{instruction\}''\\
- \textbf{Assigned Score:} \{score\} / 4\\
- \textbf{Images:} [Original Image], [Edited Image]

\textbf{**Job Description:**}

1. \textbf{Identify Edit Regions (Grounding):}
\begin{itemize}[leftmargin=*, itemsep=0pt, topsep=2pt]
\item Analyze the instruction and the changes in the images to identify \textbf{ALL distinct objects or regions} involved in the edit.
\item For each distinct edit target (up to k regions), create a bounding box [ymin, xmin, ymax, xmax] (0-1000 scale) and assign it a unique ID starting from 0.
\item Example: If the instruction is ``Change the cat to a dog and make the grass blue'', identify Region 0 (dog) and Region 1 (grass).
\end{itemize}

2. \textbf{Draft the Reasoning:}
\begin{itemize}[leftmargin=*, itemsep=0pt, topsep=2pt]
\item Write a detailed justification for the score of \{score\}.
\item \textbf{Mandatory Tagging:} You MUST reference specific regions using their tags \texttt{<|bbox\_0|>}, \texttt{<|bbox\_1|>}, ..., \texttt{<|bbox\_k|>} immediately before discussing them.
\item Use \texttt{<|global|>} to discuss the overall context, background preservation, and composition.
\end{itemize}

3. \textbf{Align with Score \{score\}:}
\begin{itemize}[leftmargin=*, itemsep=0pt, topsep=2pt]
\item \textbf{Score 4:} All \texttt{<|bbox\_k|>} regions are edited perfectly. \texttt{<|global|>} background is perfectly preserved.
\item \textbf{Score 3:} Main goal achieved, but one specific \texttt{<|bbox\_k|>} has a minor flaw (e.g., texture/color off), or \texttt{<|global|>} has slight noise.
\item \textbf{Score 2:} A major \texttt{<|bbox\_k|>} is misinterpreted/failed, or \texttt{<|global|>} background is significantly corrupted/over-edited.
\item \textbf{Score 1:} Complete failure to follow instruction or image collapse.
\end{itemize}

\textbf{**Output Format (JSON Only):}
\begin{verbatim}
{
    "edit_region": [
        {"id": 0, "label": "concise_label_for_first_target",
         "bbox_2d": [ymin, xmin, ymax, xmax]},
        {"id": 1, "label": "concise_label_for_second_target",
         "bbox_2d": [ymin, xmin, ymax, xmax]}
    ],
    "reasoning": "<|bbox_0|> [Detailed analysis of region 0]
                  <|bbox_1|> [Detailed analysis of region 1] ...
                  <|global|> [Analysis of unedited regions and
                  global consistency]"
}
\end{verbatim}
\end{promptbox}

\subsection{Visual Quality Evaluation Prompt}

\begin{promptbox}{VISUAL QUALITY DATA CONSTRUCTION TEMPLATE}
\small
You are an expert visual quality inspector.
Your task is to generate a detailed critique of an AI-generated image based on a pre-assigned quality score.

\textbf{**Inputs:**}\\
- \textbf{Assigned Score:} \{score\} / 4\\
- \textbf{Image:} [Edited Image]

\textbf{**Job Description:}\\
Analyze the image strictly for \textbf{Visual Quality} (Physics, Lighting, Artifacts) to justify the score of \{score\}.
\begin{itemize}[leftmargin=*, itemsep=0pt, topsep=2pt]
\item Do NOT discuss the text instruction.
\item \textbf{Score 4 details:} Mention perfect lighting, realistic textures, and no artifacts.
\item \textbf{Score 3 details:} Mention high quality but verify a specific minor flaw (e.g., ``slight noise in shadow'').
\item \textbf{Score 2 details:} Point out obvious flaws like ``distorted face,'' ``blurred edges,'' or ``inconsistent shadows.''
\item \textbf{Score 1 details:} Describe severe failure (garbage output).
\end{itemize}

\textbf{**Output Format (JSON Only):}
\begin{verbatim}
{
    "reasoning": "[Detailed reasoning text describing
                   the style, lighting, and any artifacts.
                   Be specific about why it receives
                   a score of {score}.]"
}
\end{verbatim}
\end{promptbox}

\subsection{Caption-Only Control Prompt}

\begin{promptbox}{CAPTION-ONLY CONTROL DATA CONSTRUCTION TEMPLATE}
\small
You are an image-pair captioning assistant.
Your task is to write a short descriptive caption for an image editing pair. The caption should describe the visible content and the main change between the original image and the edited image. Do not provide a reward score, do not explain whether the edit is good or bad, and do not write an evaluative critique.

\textbf{**Inputs:**}\\
- \textbf{Instruction:} ``\{instruction\}''\\
- \textbf{Images:} [Original Image], [Edited Image]

\textbf{**Job Description:**}
\begin{itemize}[leftmargin=*, itemsep=0pt, topsep=2pt]
\item Briefly describe the original image.
\item Briefly describe the edited image.
\item Mention the main visible change related to the instruction.
\item Avoid judging quality, instruction-following correctness, or assigning any score.
\end{itemize}

\textbf{**Output Format (JSON Only):}
\begin{verbatim}
{
    "caption": "[A short neutral description of the image pair
                and the visible edit, without reward evaluation.]"
}
\end{verbatim}
\end{promptbox}

\subsection{Score-Unconditioned Explanation Control Prompt}

\begin{promptbox}{SCORE-UNCONDITIONED CONTROL DATA CONSTRUCTION TEMPLATE}
\small
You are an expert AI training data annotator and visual editing evaluator.
Your task is to generate a structured critique for an image editing pair. You are not given the assigned reward score. Do not infer, mention, or output a numeric score. The critique should be based only on the original image, the edited image, and the editing instruction.

\textbf{**Inputs:**}\\
- \textbf{Instruction:} ``\{instruction\}''\\
- \textbf{Images:} [Original Image], [Edited Image]

\textbf{**Job Description:**}

1. \textbf{Identify Edit Regions (Grounding):}
\begin{itemize}[leftmargin=*, itemsep=0pt, topsep=2pt]
\item Analyze the instruction and the image changes to identify the main edited objects or regions.
\item For each edit target, create a bounding box [ymin, xmin, ymax, xmax] (0-1000 scale) and assign it a unique ID starting from 0.
\end{itemize}

2. \textbf{Write a Score-Free Critique:}
\begin{itemize}[leftmargin=*, itemsep=0pt, topsep=2pt]
\item Use \texttt{<|bbox\_0|>}, \texttt{<|bbox\_1|>}, ..., \texttt{<|bbox\_k|>} before discussing specific edited regions.
\item Use \texttt{<|global|>} to discuss background preservation, global consistency, artifacts, lighting, and realism.
\item Describe successes and failures based on visual evidence.
\item Do not use any pre-assigned score, and do not organize the explanation around a known score.
\end{itemize}

\textbf{**Output Format (JSON Only):}
\begin{verbatim}
{
    "edit_region": [
        {"id": 0, "label": "concise_label_for_first_target",
         "bbox_2d": [ymin, xmin, ymax, xmax]}
    ],
    "reasoning": "<|bbox_0|> [Evidence-based critique of region 0]
                  <|global|> [Analysis of background, consistency,
                  artifacts, and overall editing quality]"
}
\end{verbatim}
\end{promptbox}

\subsection{Control Variant Construction}
\label{appendix:control_variants}

For the language-supervision controls in Table~\ref{tab:language_controls}, all variants are built from the same 200K EditReward training samples and use the same reward-model architecture. Each sample keeps the original image, edited image, editing instruction, and preference or score annotation. The main JRM corpus uses GPT-5 to generate explanation-style targets with the prompts above, keeping a one-to-one correspondence with the original reward-learning samples.

We construct three control variants. The caption-only variant rewrites each sample into a short image-pair description without evaluative critique. The mismatched-explanation variant randomly shuffles the generated explanations across training samples, preserving the text pool while breaking sample-text alignment. The score-unconditioned variant generates critiques from only the images and editing instruction, without providing the assigned score. Together, these variants isolate the effects of descriptive language, sample-aligned explanations, and score conditioning.

\section{Training and Implementation Details}
\label{appendix:training}

\subsection{Joint Reward Model Training}

\textbf{Training Hyperparameters.} JRM is fine-tuned from the Qwen3-VL-8B-Instruct model. All experiments are conducted using PyTorch 2.5.1 and Transformers 4.56.1.

\begin{itemize}[leftmargin=*, itemsep=0pt, parsep=0pt, topsep=4pt]
    \item \textbf{Optimizer:} AdamW ($\beta_1 = 0.9$, $\beta_2 = 0.95$)
    \item \textbf{Weight Decay:} 0.1
    \item \textbf{Peak Learning Rate:} $2 \times 10^{-6}$
    \item \textbf{Learning Rate Schedule:} Cosine decay with 5\% warmup ratio
    \item \textbf{Training Epochs:} 10
    \item \textbf{Global Batch Size:} 64 (per-device batch size 2, gradient accumulation 4, across 8 GPUs)
    \item \textbf{Maximum Sequence Length:} 8192
    \item \textbf{Image Resolution:} Adaptive (200,704 pixels, equivalent to $256 \times 28 \times 28$)
    \item \textbf{Joint Loss Weight $\alpha$:} 0.7
    \item \textbf{Precision:} Mixed-precision (bfloat16)
    \item \textbf{Gradient Checkpointing:} Enabled (non-reentrant mode)
    \item \textbf{Distributed Training:} DeepSpeed ZeRO-2
\end{itemize}

\textbf{Model Architecture.} The reward model adopts a multi-head architecture with a RankNet-style discriminative head for preference ranking and a conditional language generation head for semantic explanation. We employ a learnable special token appended to the input sequence, whose final hidden state serves as the shared representation $\mathbf{h}$. The reward head outputs 2-dimensional scores for instruction following and visual quality.

\textbf{Compute Resources.} Training is conducted on 8 NVIDIA H800 (80GB) GPUs. The full training process takes approximately 40-50 GPU hours.

\begin{figure}[ht]
\centering
\includegraphics[width=0.55\textwidth]{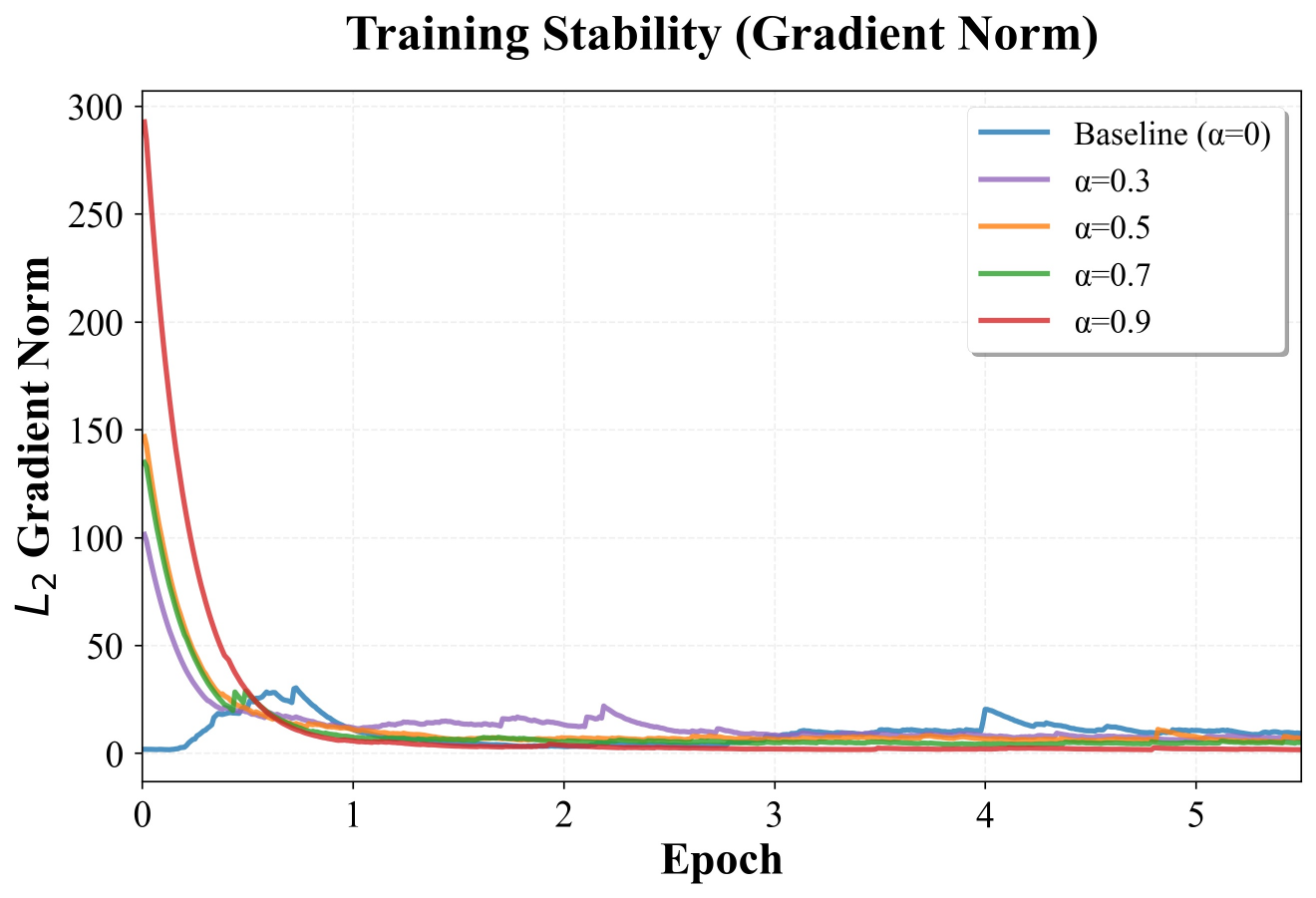}
\caption{\textbf{Gradient norm stability during JRM training.} Gradient norms under different language-supervision weights $\alpha$.}
\label{fig:grad_norm}
\end{figure}

\textbf{Training Dynamics.} Figure~\ref{fig:grad_norm} illustrates the gradient norm stability across different language supervision weights $\alpha$. Joint training with language supervision introduces larger initial gradient norms, especially for higher $\alpha$ values, but all variants rapidly stabilize within the first epoch.

\begin{figure}[ht]
\centering
\includegraphics[width=0.55\textwidth]{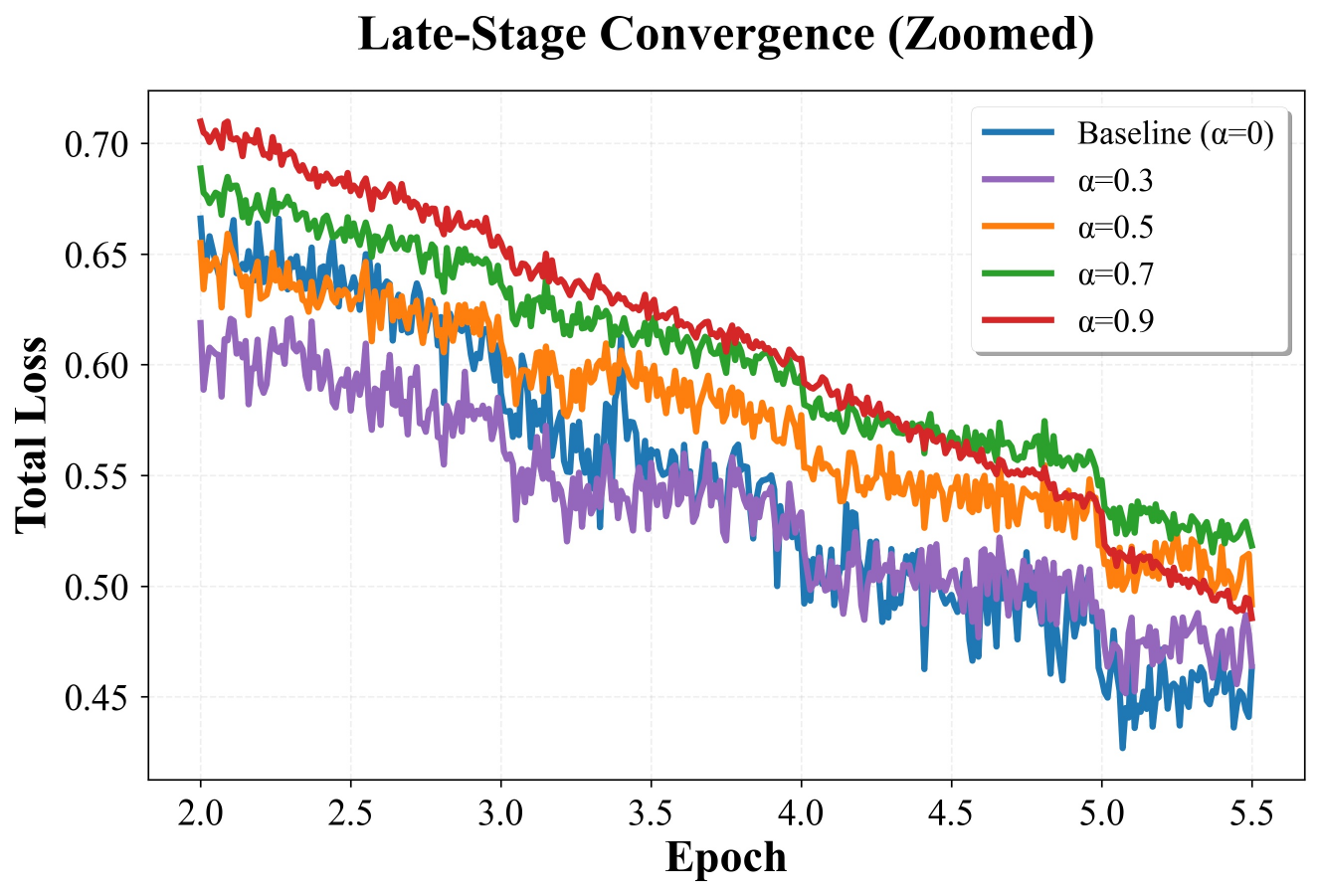}
\caption{\textbf{Late-stage loss convergence (epochs 2--5.5).} Total loss curves under different language-supervision weights $\alpha$.}
\label{fig:loss_zoom}
\end{figure}

Figure~\ref{fig:loss_zoom} shows the late-stage convergence behavior of the total loss. Higher $\alpha$ values yield larger absolute losses because of the additional language modeling objective, but all configurations converge stably.

\subsection{Reinforcement Learning Fine-Tuning}

\textbf{Training Hyperparameters.} For online RL fine-tuning of OmniGen2-Edit, we use the Flow-GRPO algorithm.

\begin{itemize}[leftmargin=*, itemsep=0pt, parsep=0pt, topsep=4pt]
    \item \textbf{Algorithm:} Flow-GRPO (Group Relative Policy Optimization)
    \item \textbf{Discrete Timesteps $T$:} 20
    \item \textbf{Diffusion Coefficient $\sigma$:} 0.9
    \item \textbf{Global Batch Size:} 288
    \item \textbf{Group Size $G$:} 12
    \item \textbf{PPO Clipping $\epsilon_{\text{low}}$:} $10^{-4}$
    \item \textbf{PPO Clipping $\epsilon_{\text{high}}$:} $5 \times 10^{-4}$
    \item \textbf{Learning Rate:} $4 \times 10^{-4}$
    \item \textbf{KL Penalty Coefficient $\beta$:} 0.04
    \item \textbf{LoRA Configuration:} $r = 32$, $\alpha = 64$
\end{itemize}

\begin{figure}[ht]
\centering
\includegraphics[width=0.6\textwidth]{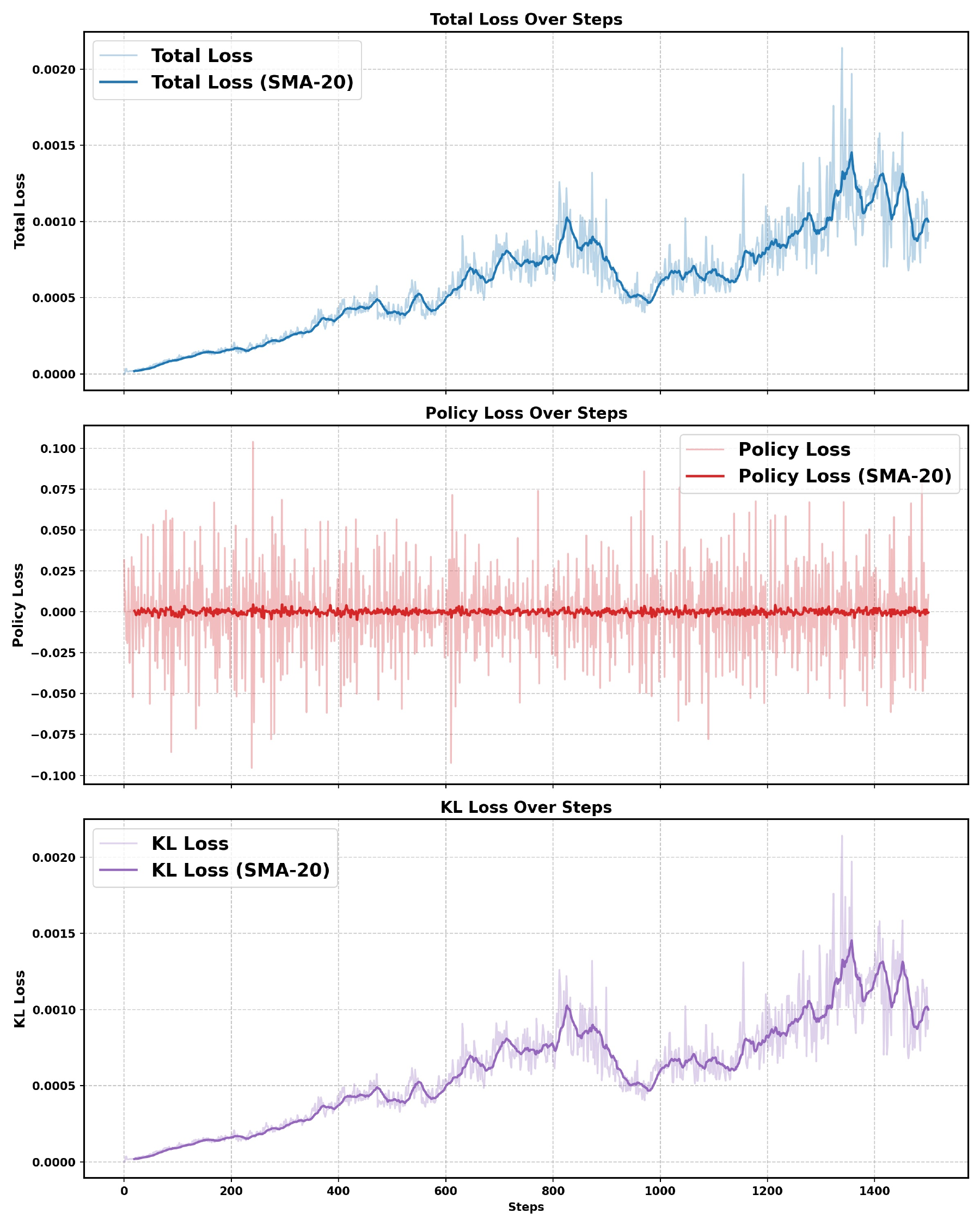}
\caption{\textbf{Loss components during Flow-GRPO training.} The figure shows the dynamics of individual loss terms during online RL fine-tuning with JRM as the reward signal.}
\label{fig:rl_loss_components}
\end{figure}

\textbf{Compute Resources.} Online RL fine-tuning is performed on 32 NVIDIA H800 (80GB) GPUs with iterative sampling and policy updates. The detailed loss dynamics during training are illustrated in Figure~\ref{fig:rl_loss_components}.

\begin{figure}[ht]
\centering
\includegraphics[width=0.7\textwidth]{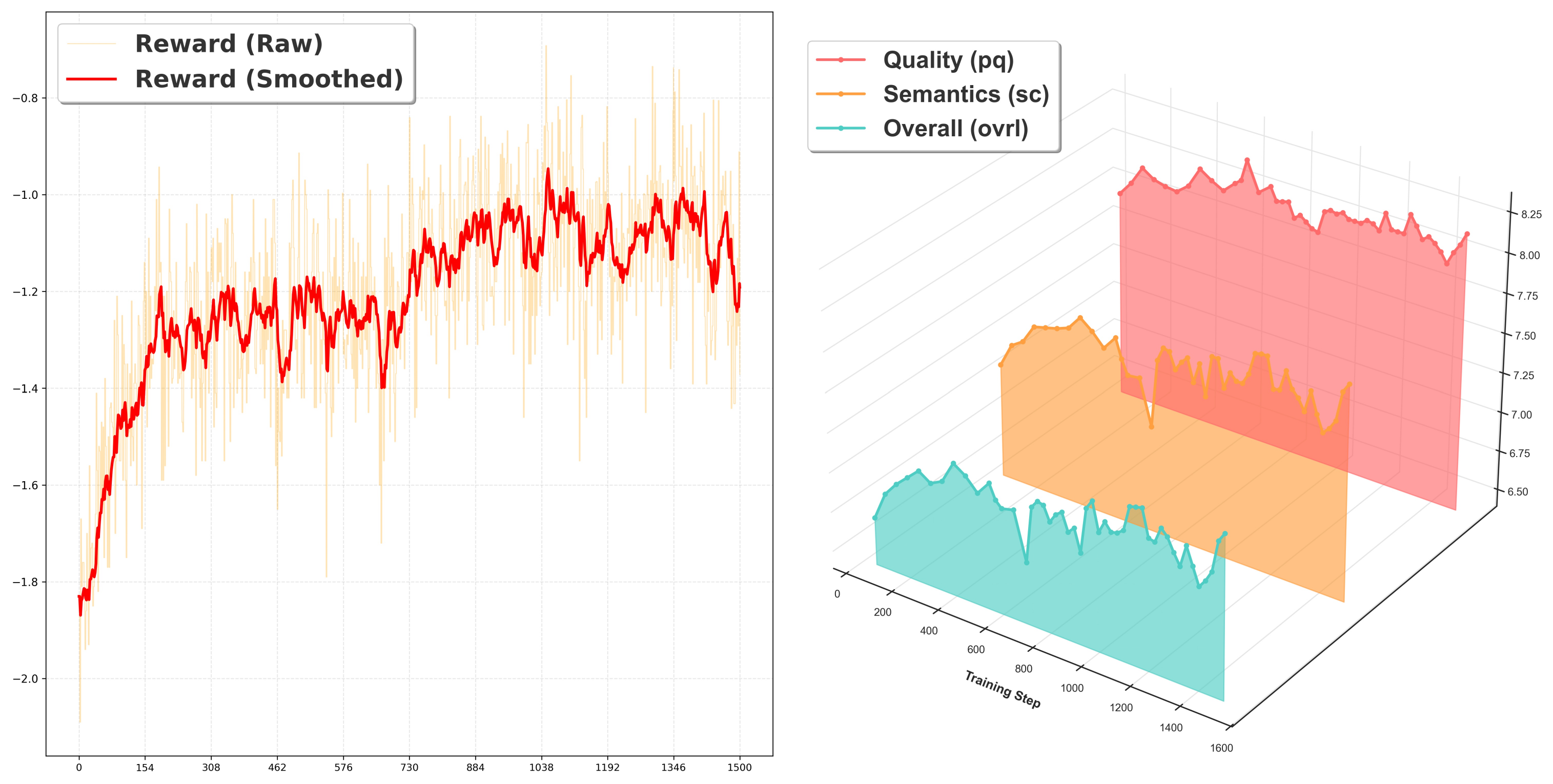}
\caption{\textbf{Training dynamics and benchmark evaluation during online RL.} \textbf{Left:} Reward curves during Flow-GRPO training with JRM. \textbf{Right:} GEdit-Bench scores across training checkpoints.}
\label{fig:rl_training}
\end{figure}

\subsection{Evaluation Protocol and Statistical Reporting}
\label{appendix:statistical}

For benchmark evaluation, each model is evaluated with five different inference seeds, and the reported benchmark scores are averaged over these five runs. This protocol reduces variance caused by stochastic sampling during image generation and provides a more stable estimate of reward-model and downstream RL performance. Due to the high cost of online RL fine-tuning, we report averaged benchmark scores rather than full confidence intervals.

To further assess the stability of downstream online RL, we additionally repeat JRM-guided RL fine-tuning with three independent RL seeds. Table~\ref{tab:rl_seed_stability} reports the mean and standard deviation of the improvement over the base OmniGen2 model. The low standard deviations indicate that the downstream gains are stable across RL runs.

\begin{table}[ht]
    \centering
    \footnotesize
    \setlength{\tabcolsep}{5pt}
    \caption{\textbf{Seed stability of downstream RL gains.} Mean $\pm$ standard deviation over three independent online RL seeds.}
    \label{tab:rl_seed_stability}
    \begin{tabular}{l c c}
        \toprule
        \textbf{Reward Signal} & \textbf{GEdit Gain $\Delta$} & \textbf{ImageEdit Gain $\Delta$} \\
        \midrule
        JRM & $+0.95 \pm 0.06$ & $+0.48 \pm 0.04$ \\
        \bottomrule
    \end{tabular}
\end{table}

\subsection{Language-Weight Ablation Values}
\label{appendix:alpha_ablation_values}

\begin{table}[ht]
    \centering
    \footnotesize
    \setlength{\tabcolsep}{8pt}
    \caption{\textbf{Exact values for Figure~\ref{fig:ablation_training}(a).} Reward-model benchmark results under different language-supervision weights $\alpha$.}
    \label{tab:alpha_ablation_values}
    \begin{tabular}{c c c}
        \toprule
        \textbf{$\alpha$} & \textbf{EditReward-Bench} & \textbf{MMRB2} \\
        \midrule
        0.0 & 0.792 & 0.652 \\
        0.3 & 0.832 & 0.683 \\
        0.5 & 0.843 & 0.689 \\
        0.7 & 0.851 & 0.693 \\
        0.9 & 0.847 & 0.709 \\
        \bottomrule
    \end{tabular}
\end{table}

\subsection{Matched-Backbone Comparison}
\label{appendix:matched_backbone}

To reduce the confound from different backbone models, we additionally train JRM with Qwen2.5-VL-7B and compare it with Qwen2.5-VL-7B results reported by EditScore and EditReward. Only the JRM row is our retraining.

\begin{table}[ht]
    \centering
    \footnotesize
    \setlength{\tabcolsep}{4pt}
    \caption{\textbf{Matched-backbone comparison on Qwen2.5-VL-7B.} JRM remains stronger than the reported Qwen2.5-VL-7B baselines.}
    \label{tab:matched_backbone}
    \resizebox{\textwidth}{!}{%
    \begin{tabular}{l l l c c c c}
        \toprule
        \textbf{Model} & \textbf{Backbone} & \textbf{Source} & \textbf{EditReward-Bench} & \textbf{MMRB2} & \textbf{GEdit $\Delta$} & \textbf{ImageEdit $\Delta$} \\
        \midrule
        JRM & Qwen2.5-VL-7B & Our retraining & 0.837 & 0.684 & +0.92 & +0.37 \\
        EditReward~\cite{editreward} & Qwen2.5-VL-7B & Reported & 0.792 & 0.657 & +0.77 & +0.19 \\
        EditScore avg@4~\cite{editscore} & Qwen2.5-VL-7B & Reported & 0.722 & 0.608 & +0.40 & +0.23 \\
        \bottomrule
    \end{tabular}}
\end{table}

\section{JRM Inference Prompts}
\label{appendix:inference}

Standard JRM inference uses only the discriminative reward head. The language head can be queried for optional diagnostic explanations, and the corresponding prompt templates are provided below.

\subsection{Instruction Following Inference Prompt}

\begin{promptbox}{INSTRUCTION EDIT FOLLOWING TEMPLATE}
\small
You are tasked with evaluating an edited image \textbf{in comparison with the original source image} based on \textbf{Instruction Following \& Semantic Fidelity}, and assigning a score from 1 to 4, with 1 being the worst and 4 being the best.
This dimension focuses on how accurately, completely, and exclusively the model executed the given text instruction.

\textbf{**Inputs Provided:}\\
- Source Image (before editing)\\
- Edited Image (after applying the instruction)\\
- Text Instruction

\textbf{**Sub-Dimensions to Evaluate:}
\begin{itemize}[leftmargin=*, itemsep=0pt, topsep=2pt]
\item \textbf{Semantic Accuracy:} Assess whether the edited content accurately captures the semantics of the instruction. The edited result should precisely match the intended meaning. For example, if the instruction is ``replace apples with oranges,'' the object must clearly be oranges, not other fruits.
\item \textbf{Completeness of Editing:} Check whether \textbf{all parts} of the instruction are fully executed. For multi-step edits (e.g., ``replace a red car with a blue bicycle''), both the color change and the object replacement must be done without omissions.
\item \textbf{Exclusivity of Edit (No Over-Editing):} Ensure that only the requested parts are changed. The rest of the image (as seen in the source) should remain unaltered. For example, if the instruction only involves replacing an object, the background, lighting, and unrelated objects should not be unnecessarily modified.
\end{itemize}

\textbf{**Scoring Criteria:}
\begin{itemize}[leftmargin=*, itemsep=0pt, topsep=2pt]
\item \textbf{4 (Very Good):} Perfectly accurate, complete, and exclusive execution of the instruction.
\item \textbf{3 (Relatively Good):} Largely correct, but with minor omissions or slight over-editing.
\item \textbf{2 (Relatively Poor):} Major misinterpretation, incomplete edits, or noticeable unintended changes.
\item \textbf{1 (Very Poor):} Instruction ignored or completely wrong execution.
\end{itemize}

\textbf{**IMPORTANT: Output Format}

You must provide your output in this format:
\begin{verbatim}
{"edit_region": [...], "reasoning": "..."}
\end{verbatim}

First, identify where the editing occurred in the second image:
\begin{itemize}[leftmargin=*, itemsep=0pt, topsep=2pt]
\item If editing was successful, provide bounding boxes with labels: \texttt{[\{"id": 0$\sim$n, "label": "description of edited area", "bbox\_2d": [x1, y1, x2, y2]\}]} (coordinates normalized to [0, 1000] range)
\item If editing failed (images look identical), use empty list: \texttt{[]}
\end{itemize}

In your reasoning, use special tokens to reference regions:
\begin{itemize}[leftmargin=*, itemsep=0pt, topsep=2pt]
\item \texttt{<|bbox\_\{id\}|>} before describing each edited region (if exist)
\item \texttt{<|global|>} before overall assessment
\end{itemize}

Text instruction - \{text\_prompt\}
\end{promptbox}

\subsection{Visual Quality Inference Prompt}

\begin{promptbox}{VISUAL QUALITY EVALUATION TEMPLATE}
\small
You are tasked with evaluating an edited image \textbf{in comparison with the original source image} based on \textbf{Visual Quality \& Realism}, and assigning a score from 1 to 4, with 1 being the worst and 4 being the best.
This dimension focuses on how realistic, artifact-free, and aesthetically appealing the edited image is, while remaining consistent with the source image.

\textbf{**Inputs Provided:}\\
- Source Image (before editing)\\
- Edited Image (after applying the instruction)\\
- Text Instruction

\textbf{**Sub-Dimensions to Evaluate:}
\begin{itemize}[leftmargin=*, itemsep=0pt, topsep=2pt]
\item \textbf{Plausibility \& Physical Consistency:} Check whether the edit aligns with the laws of physics and the scene context. Lighting, shadows, reflections, perspective, size, and interactions with the environment should all appear natural compared to the source image.
\item \textbf{Artifact-Free Quality:} Look for technical flaws such as blur, distortions, pixel misalignment, unnatural textures, or seams around edited regions. High-quality results should be free from such visible artifacts.
\item \textbf{Aesthetic Quality:} Evaluate the overall harmony and visual appeal. The image should look natural, balanced, and pleasant. Colors, composition, and atmosphere should enhance the image rather than degrade it.
\end{itemize}

\textbf{**Scoring Criteria:}
\begin{itemize}[leftmargin=*, itemsep=0pt, topsep=2pt]
\item \textbf{4 (Very Good):} Perfectly realistic, artifact-free, seamless, and aesthetically pleasing.
\item \textbf{3 (Relatively Good):} Mostly realistic and clean, with only minor flaws that do not significantly distract.
\item \textbf{2 (Relatively Poor):} Noticeable physical inconsistencies or visible artifacts that make the edit unnatural.
\item \textbf{1 (Very Poor):} Severe artifacts, incoherent composition, or visually unusable result.
\end{itemize}

Text instruction - \{text\_prompt\}
\end{promptbox}

\vspace{1em}
Note that during standard discriminative inference, JRM bypasses the language head and directly outputs reward scores through the efficient discriminative pathway, ensuring low-latency evaluation suitable for online reinforcement learning.

\section{Limitations}
\label{appendix:limitations}

JRM is designed for efficient reward modeling in image editing, and the current study focuses on this setting. While the results are consistent across multiple reward benchmarks and downstream RL evaluations, future work can further examine the same joint-training idea on broader multimodal generation tasks and additional editing backbones. The method also relies on language supervision data generated from existing evaluation protocols; improving the diversity and coverage of these explanations may further strengthen the learned representations. Finally, although JRM keeps the inference path efficient by using the discriminative reward head, joint training and online RL validation still require non-trivial GPU resources. We expect these costs to decrease with more optimized training recipes and lighter backbone architectures.

\section{Broader Impact and Safeguards}
\label{appendix:responsible}

\textbf{Broader Impact.} JRM aims to improve the reliability and efficiency of reward feedback for image editing systems. Potential positive impacts include more accurate automatic evaluation, more stable online alignment, and reduced inference cost compared with reward models that require explicit reasoning generation at test time. At the same time, better reward models for image editing may also improve the optimization of systems that could be misused to create misleading edits, identity manipulation, or disinformation. These risks are inherited from image generation and editing technologies rather than introduced by the reward model alone, but they should be considered when deploying JRM-guided training pipelines.

\textbf{Safeguards.} We plan to release the JRM reward model, code, and explanatory supervision dataset for reward modeling, evaluation, and alignment research, but this work does not release a new image generator. The released model will be accompanied by usage notes discouraging deceptive, privacy-invasive, or harmful image editing applications. Any released data or examples will be filtered to avoid unsafe or privacy-sensitive content. When JRM is used in downstream RL pipelines, it should be combined with existing safety filters and human review for high-risk deployment scenarios.

\textbf{Assets and Licenses.} Existing models, datasets, and benchmarks used in this work are cited in the main text or bibliography. We follow their licenses and terms of use, and will provide the JRM model, code, explanatory supervision dataset, and data-construction documentation in the supplementary material or an anonymized release.

\section{Qualitative Comparison of RL-Fine-Tuned Models}
\label{appendix:rl_qualitative}

This section presents qualitative comparisons of image editing results generated by OmniGen2 models fine-tuned with different reward signals through reinforcement learning. The following figures show side-by-side comparisons across four conditions: the source image, the base OmniGen2 model, the OmniGen2 model fine-tuned with a baseline reward model, and the OmniGen2 model fine-tuned with JRM (Ours).

\begin{figure}[ht]
\centering
\includegraphics[width=\textwidth, page=1]{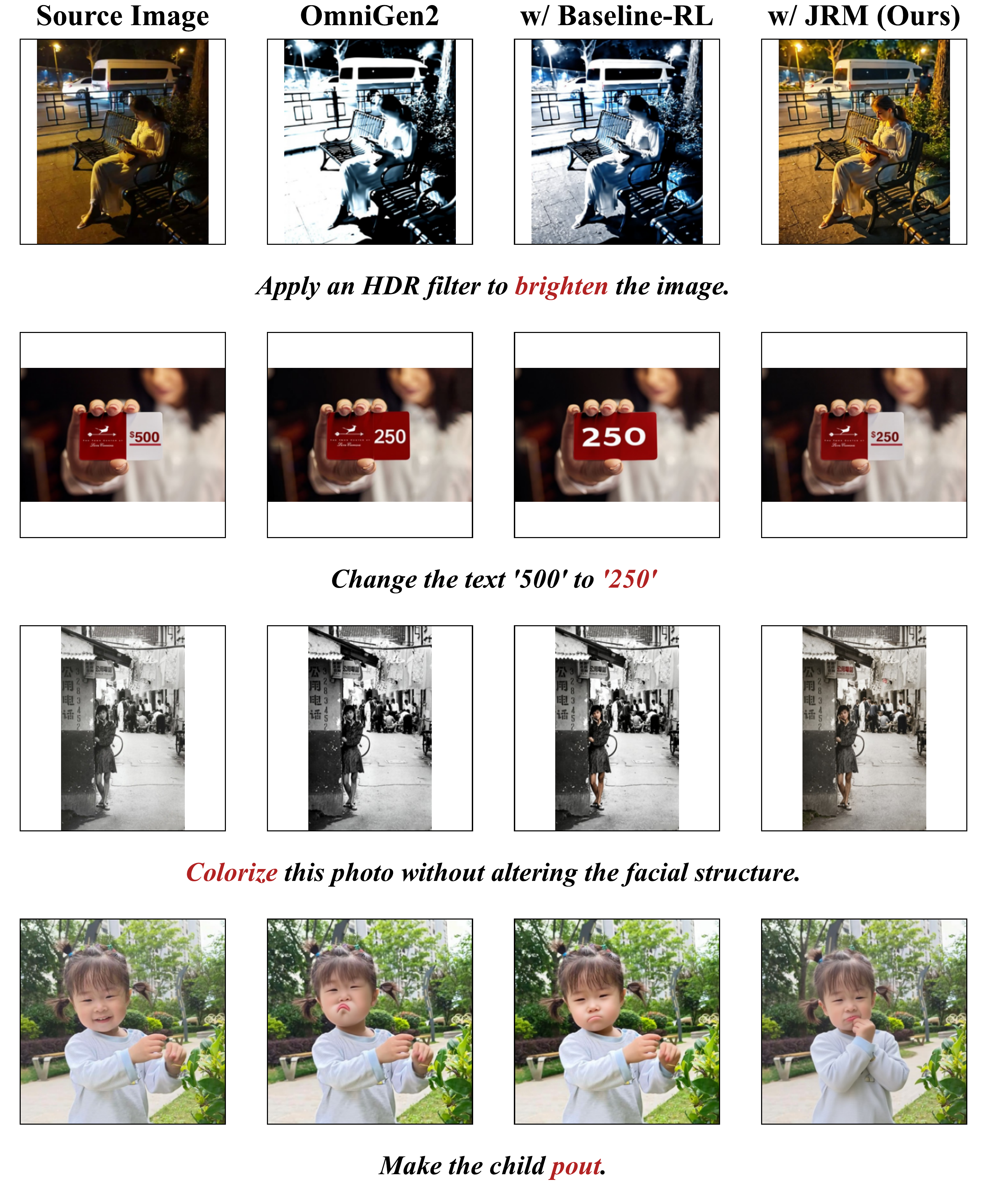}
\caption{\textbf{Qualitative comparison of RL-fine-tuned OmniGen2 models (Part 1/4).} From left to right: Source Image, base OmniGen2, OmniGen2 fine-tuned with baseline reward model (w/ Baseline-RL), and OmniGen2 fine-tuned with JRM (w/ JRM).}
\label{fig:rl_app_qualitative_1}
\end{figure}

\begin{figure}[ht]
\centering
\includegraphics[width=\textwidth, page=2]{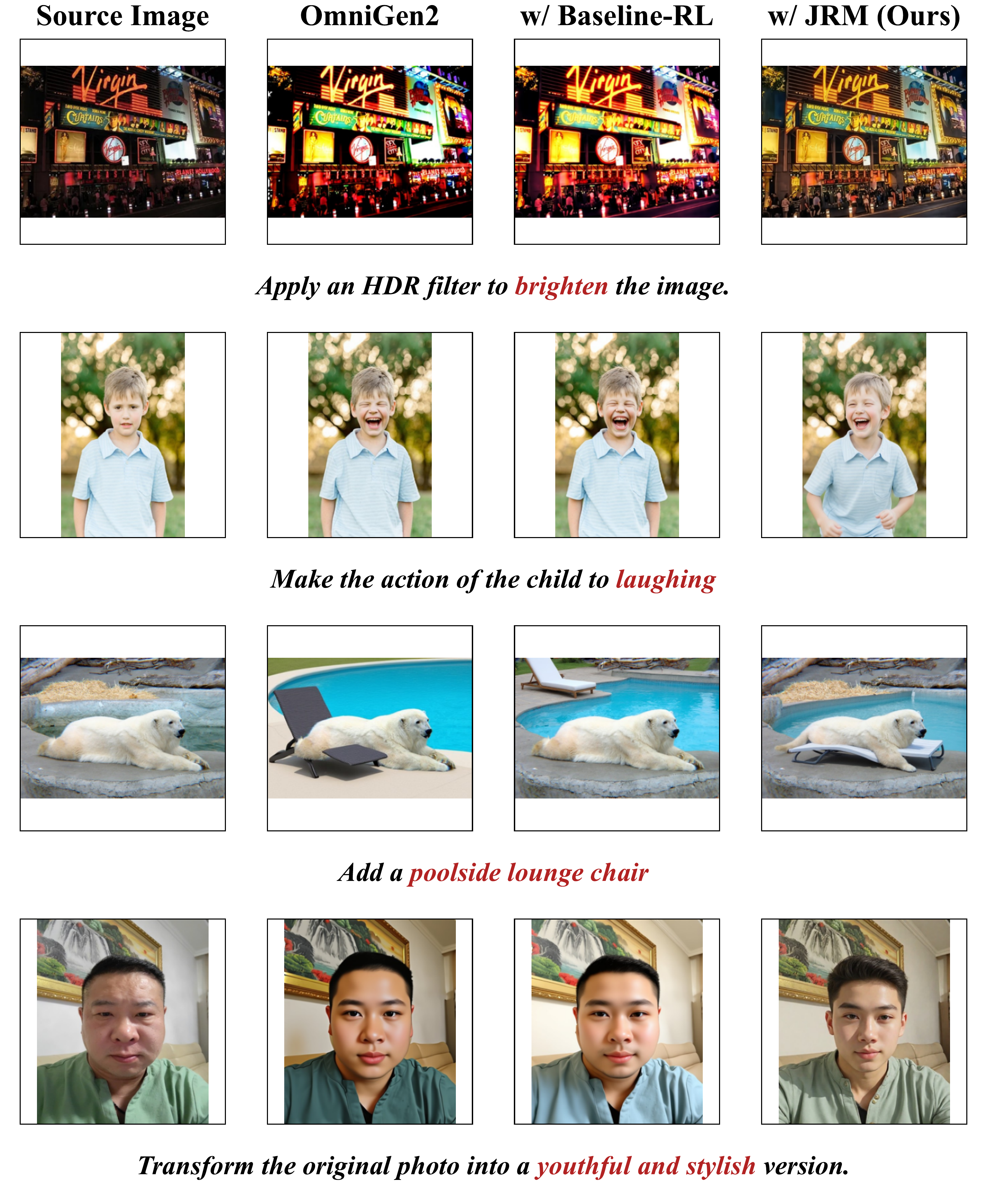}
\caption{\textbf{Qualitative comparison of RL-fine-tuned OmniGen2 models (Part 2/4).}}
\label{fig:rl_app_qualitative_2}
\end{figure}

\begin{figure}[ht]
\centering
\includegraphics[width=\textwidth, page=3]{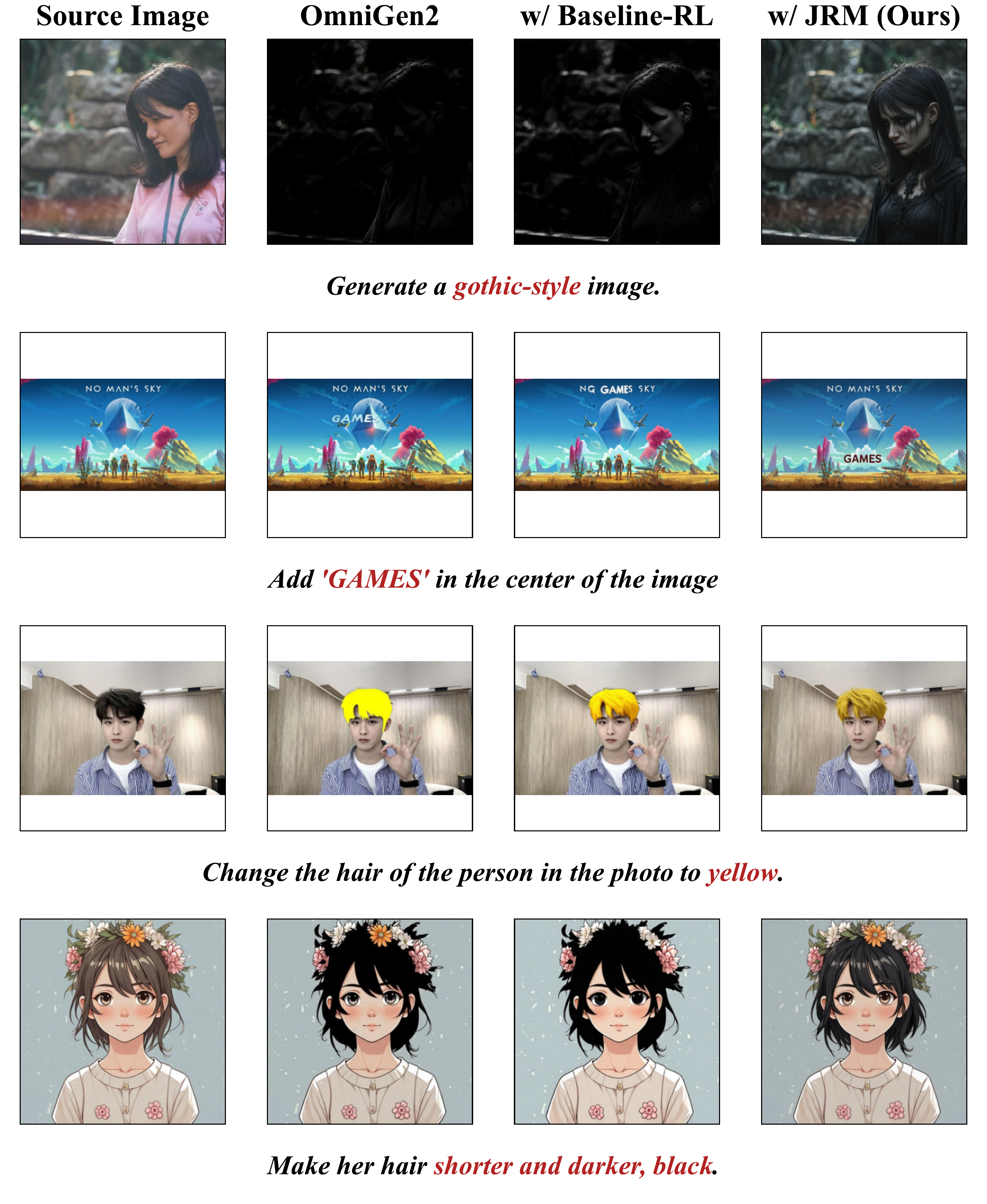}
\caption{\textbf{Qualitative comparison of RL-fine-tuned OmniGen2 models (Part 3/4).}}
\label{fig:rl_app_qualitative_3}
\end{figure}

\begin{figure}[ht]
\centering
\includegraphics[width=\textwidth, page=4]{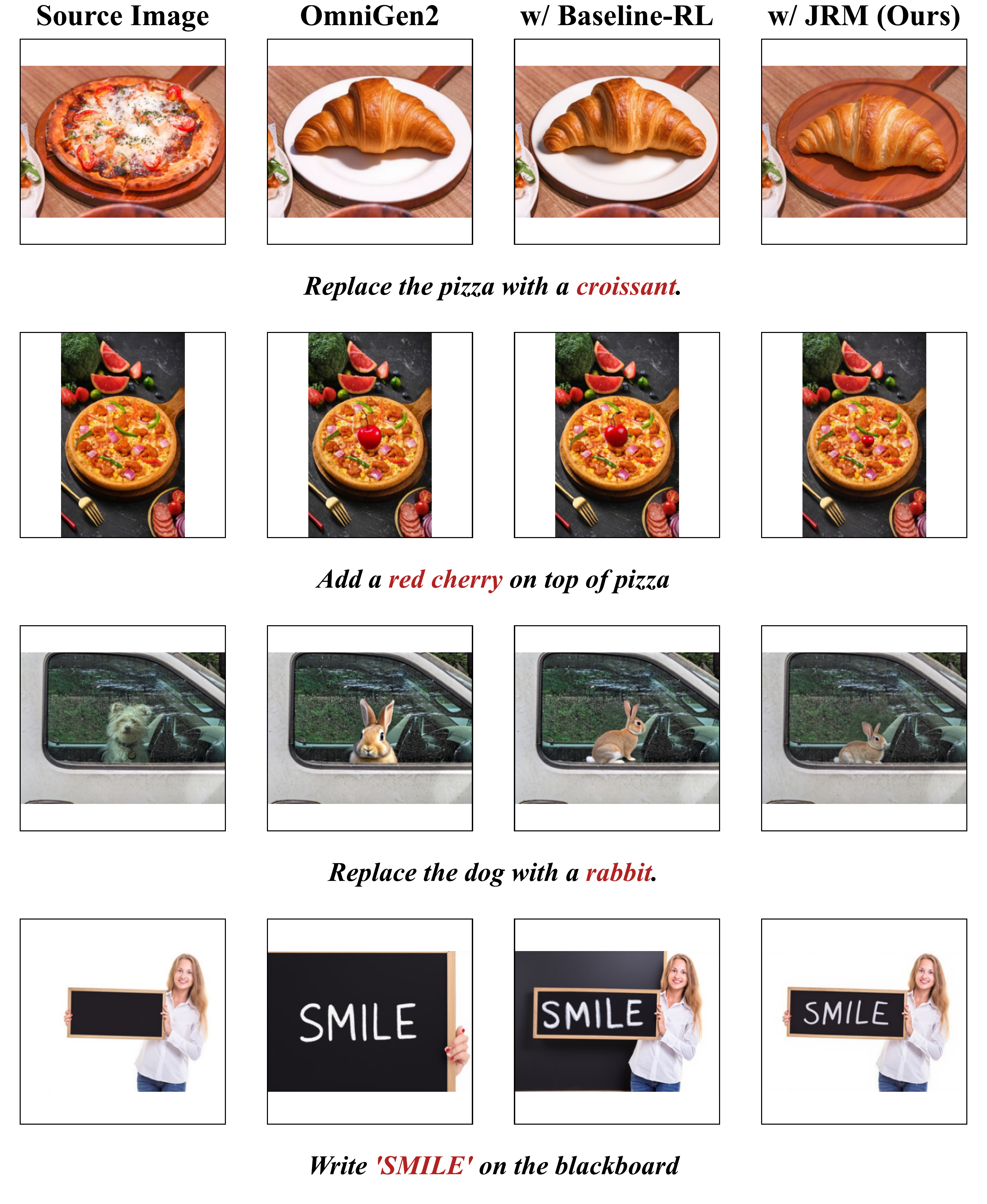}
\caption{\textbf{Qualitative comparison of RL-fine-tuned OmniGen2 models (Part 4/4).}}
\label{fig:rl_app_qualitative_4}
\end{figure}

\end{document}